\documentclass[twocolumn,10pt]{asme2e}

\usepackage{amsmath}
\usepackage{amssymb}
\usepackage{graphicx}
\usepackage{textcomp}
\usepackage{xcolor}
\usepackage{pgfplots}
\usepackage{subfigure}
\usepackage{caption}
\usepgfplotslibrary{groupplots}
\usepackage{tikz}
\usepackage{tikzscale}
\usepackage{cleveref}
\usepackage{float}
\usepackage{multirow}
\usepackage{algorithm}
\usepackage{algpseudocode} 
\usepackage{siunitx}
\usepackage{comment}
\usepackage{booktabs}
\usepackage{alphalph}
\usepackage{url}
\usepackage{siunitx}

\crefname{figure}{Fig.}{Fig.}
\Crefname{figure}{Figure}{Figures}
\crefname{equation}{}{}
\Crefname{equation}{Equation}{Equations}

\newcommand{\ie}{\textit{i}.\textit{e}., }
\newcommand{\eg}{\textit{e}.\textit{g}., }
\newcommand{\st}{\text{s.t. }}

\confshortname{International Symposium on Flexible Automation\\ ISFA 2024}
\conffullname{the ASME 2024}

\confdate{July 21-24}
\confyear{2024}
\confcity{Seattle}
\confcountry{WA}

\papernum{ISFA2024-139981}

\title{A Lightweight and Transferable Design \\ for Robust Lego Manipulation}

\author{Ruixuan Liu, Yifan Sun, Changliu Liu \thanks{This work is in part supported by Siemens and Manufacturing Futures Institute, Carnegie Mellon University, through a grant from the Richard King Mellon Foundation.} 
\thanks{Contact author: ruixuanl, yifansu2, cliu6@andrew.cmu.edu}
    \affiliation{
    Robotics Institute\\
	Carnegie Mellon University\\
	Pittsburgh, PA, USA
    }	
}

\begin{document}

\maketitle    

\begin{abstract}
{Lego is a well-known platform for prototyping pixelized objects.
However, robotic Lego prototyping (\ie manipulating Lego bricks) is challenging due to the tight connections and accuracy requirements.
This paper investigates safe and efficient robotic Lego manipulation.
In particular, this paper reduces the complexity of the manipulation by hardware-software co-design.
An end-of-arm tool (EOAT) is designed, which reduces the problem dimension and allows large industrial robots to manipulate small Lego bricks.
In addition, this paper uses evolution strategy to optimize the robot motion for Lego manipulation.
Experiments demonstrate that the EOAT can reliably manipulate Lego bricks and the learning framework can effectively and safely improve the manipulation performance to a 100\% success rate.
The co-design is deployed to multiple robots (\ie FANUC LR-mate 200id/7L and Yaskawa GP4) to demonstrate its generalizability and transferability.
In the end, we show that the proposed solution enables sustainable robotic Lego prototyping, in which the robot can repeatedly assemble and disassemble different prototypes. 
}
\end{abstract}



\section{INTRODUCTION}

With shorter life cycles of products and the rise of customization needs in manufacturing, there is a growing demand for fast automatic prototyping capabilities to meet users’ needs. 
Although 3D printing-based prototyping \cite{10.1007/978-3-319-33609-1_2,review_additive,RASIYA20216896} is mature, automatic prototyping for \textbf{assembly} remains challenging \cite{10.5555/3113190.3113359,AHMAD2015412,GOMESDESA1999389}.
In addition to constructing prototypes, \textbf{disassembly}, which is currently time-consuming and expensive \cite{9817380}, is equally critical due to environmental concerns and government regulations \cite{disassemble_article}.
Thus, it is important that the prototyping system can achieve automatic assembly as well as disassembly to ensure sustainability.

\begin{figure*}
\centering
\subfigure[Failed assembly.]{\includegraphics[width=0.24\linewidth]{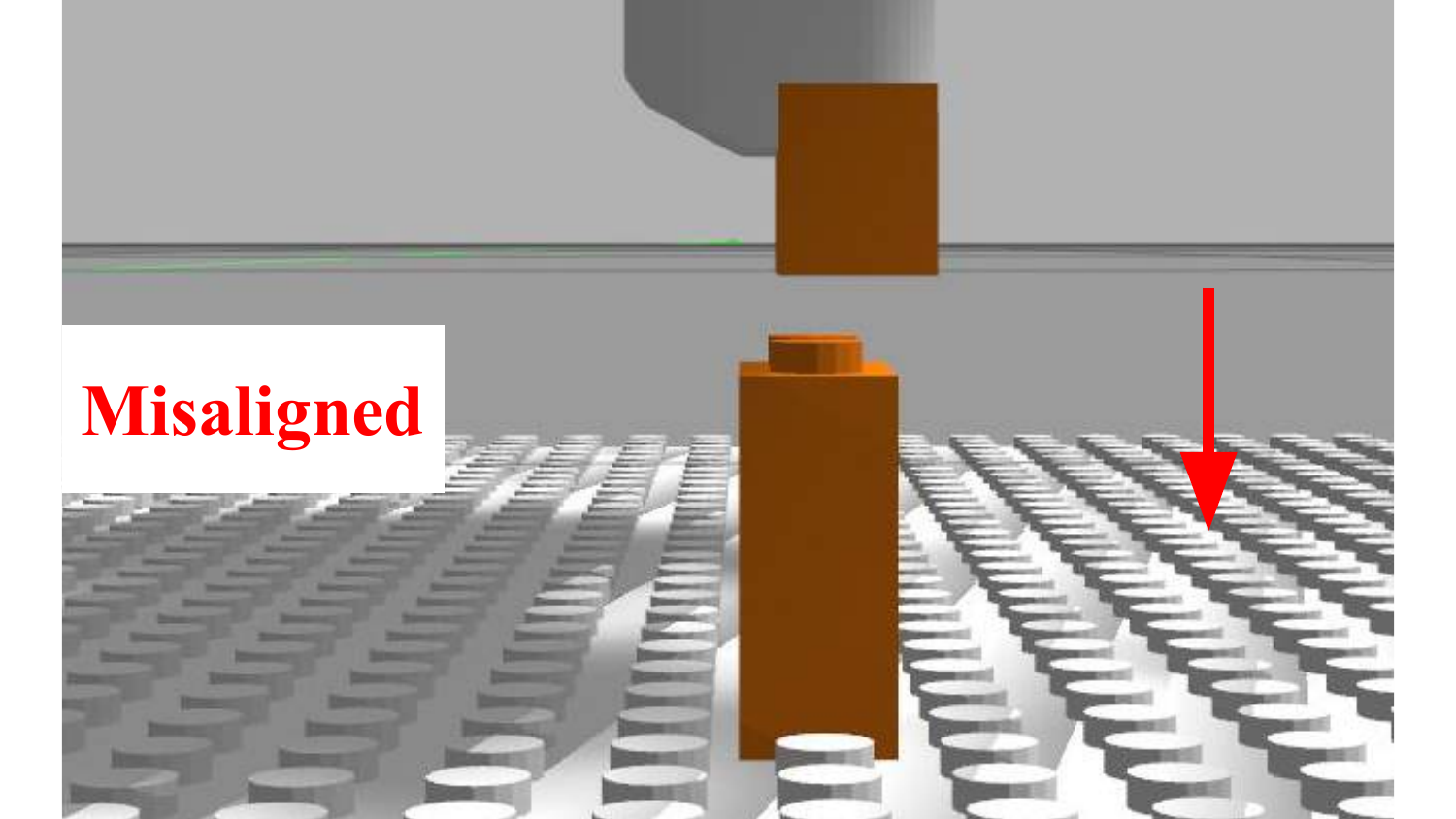}\label{fig:assemble_fail}}\hfill
\subfigure[Successful assembly.]{\includegraphics[width=0.24\linewidth]{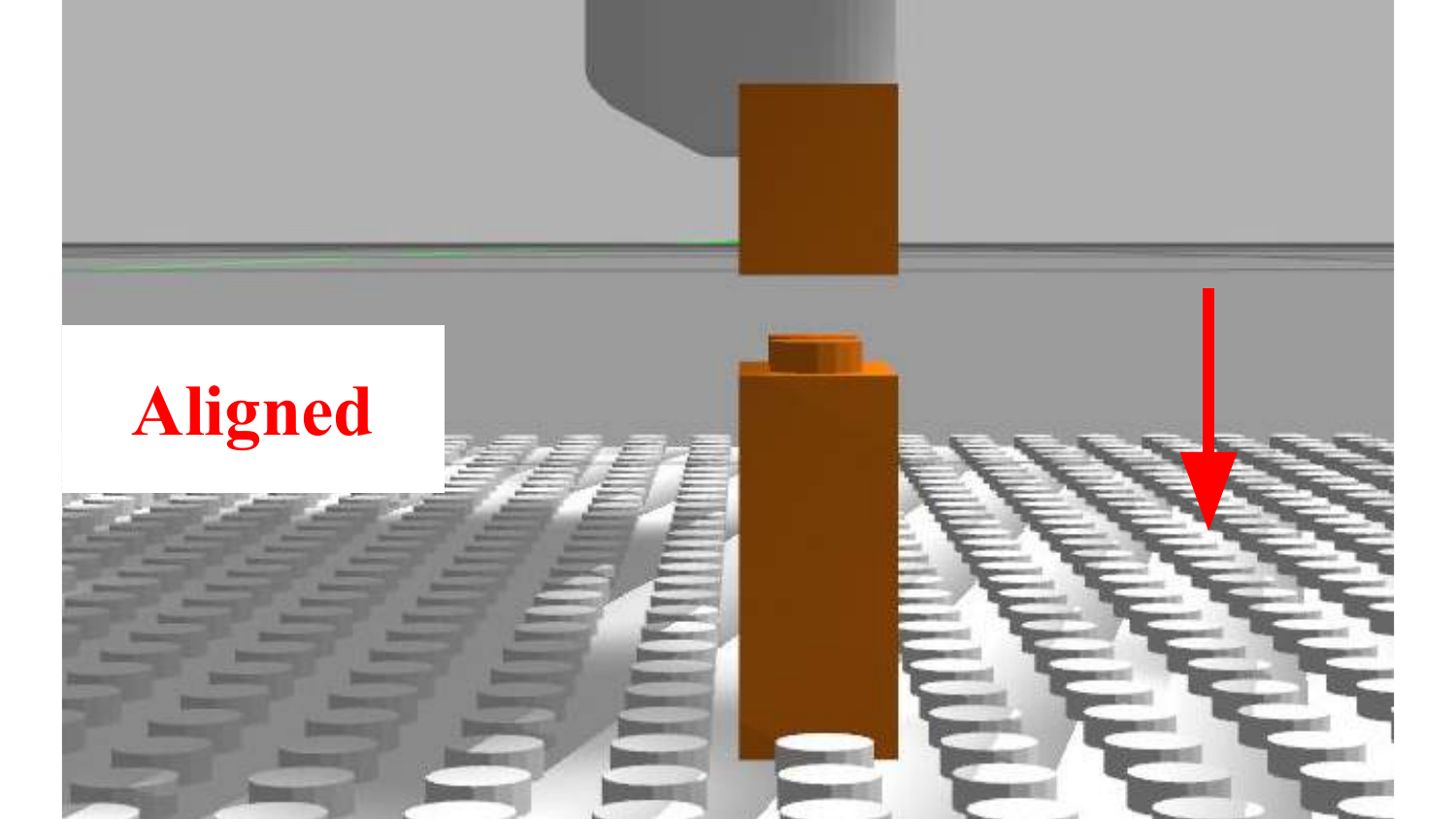}\label{fig:assemble_success}}\hfill
\subfigure[Failed disassembly.]{\includegraphics[width=0.24\linewidth]{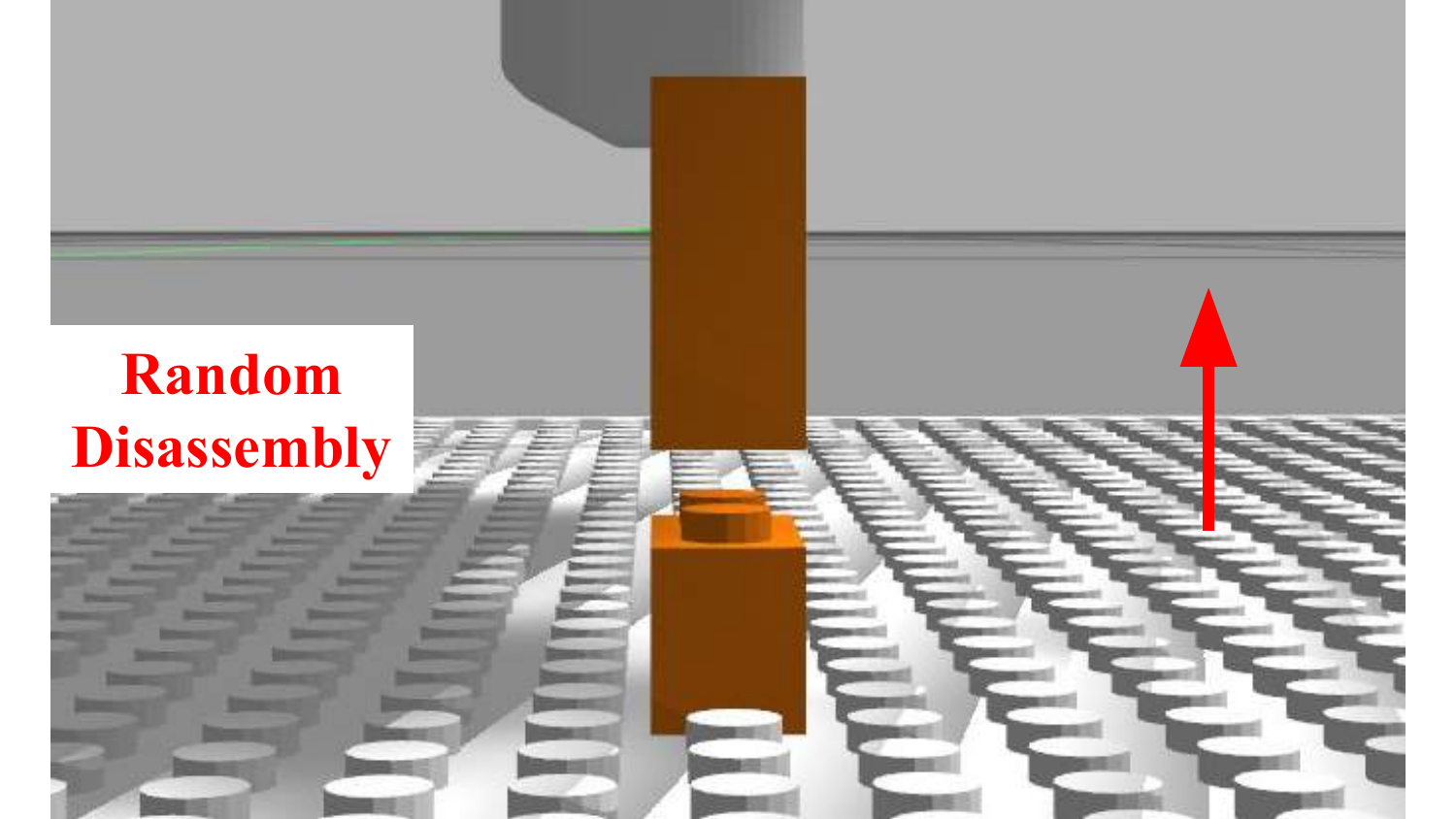}\label{fig:disassemble_fail}}\hfill
\subfigure[Successful disassembly.]{\includegraphics[width=0.24\linewidth]{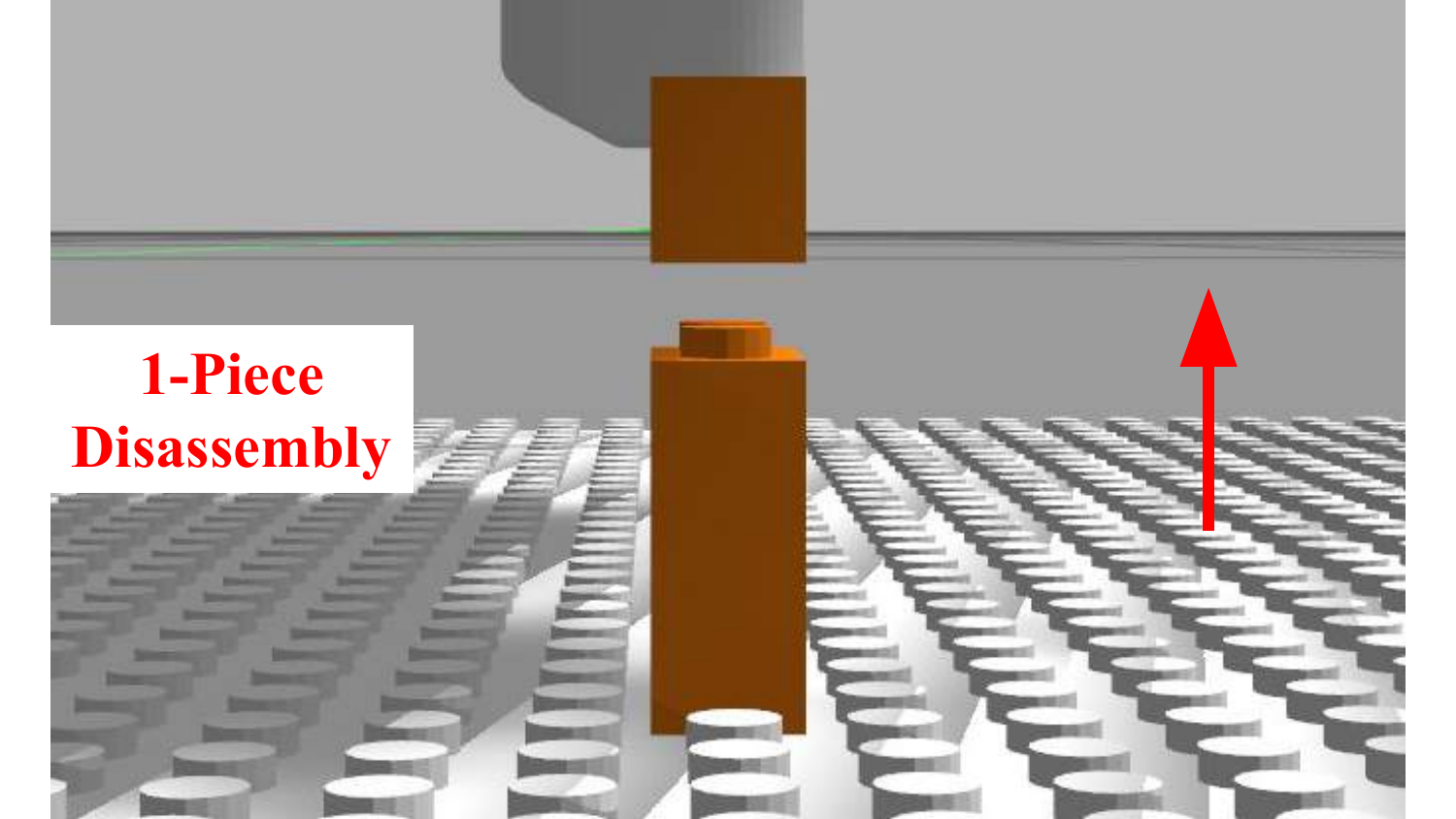}\label{fig:disassemble_success}}\\
\subfigure[Characters: AI.]{\includegraphics[width=0.16\linewidth]{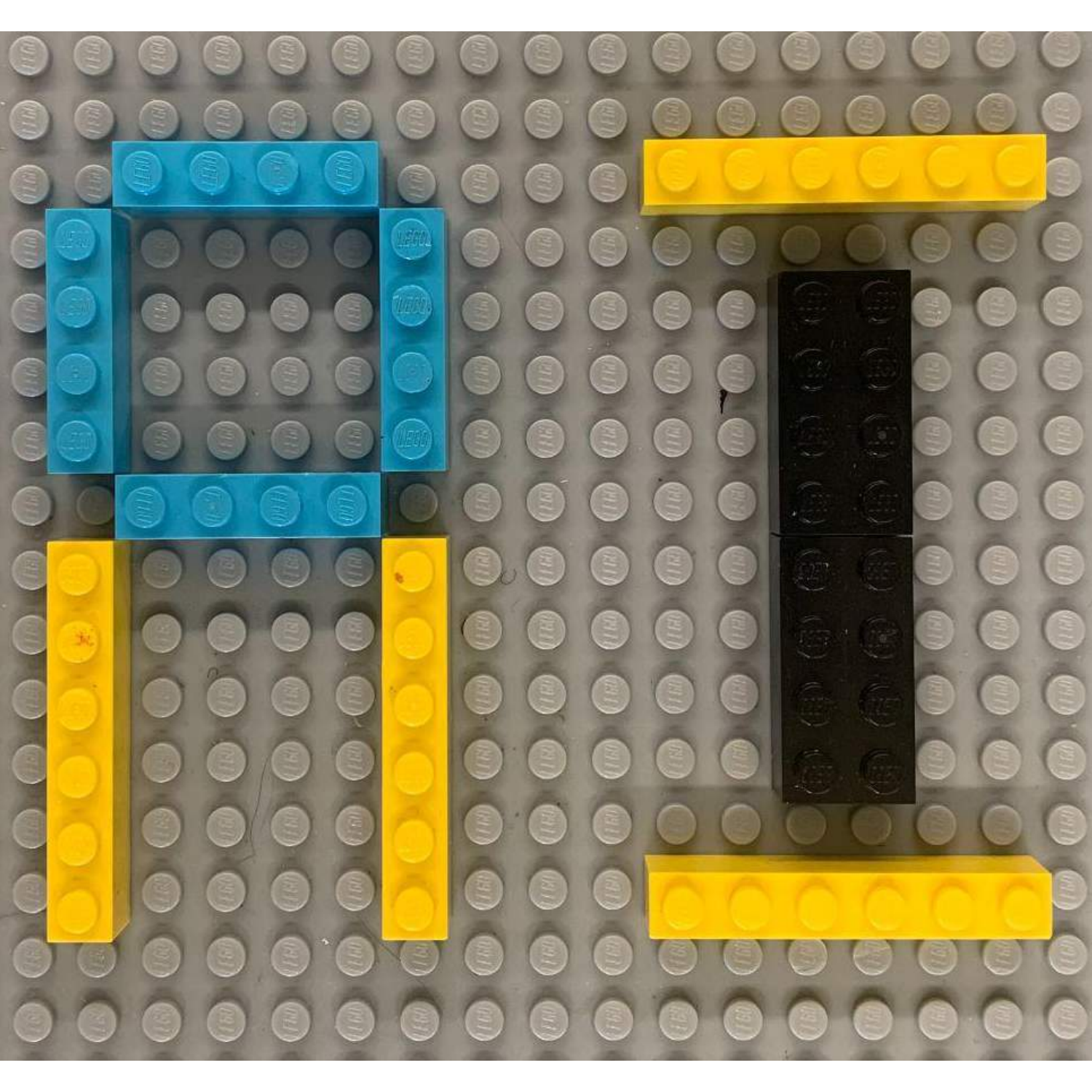}\label{fig:ai}}\hfill
\subfigure[Characters: RI.]{\includegraphics[width=0.16\linewidth]{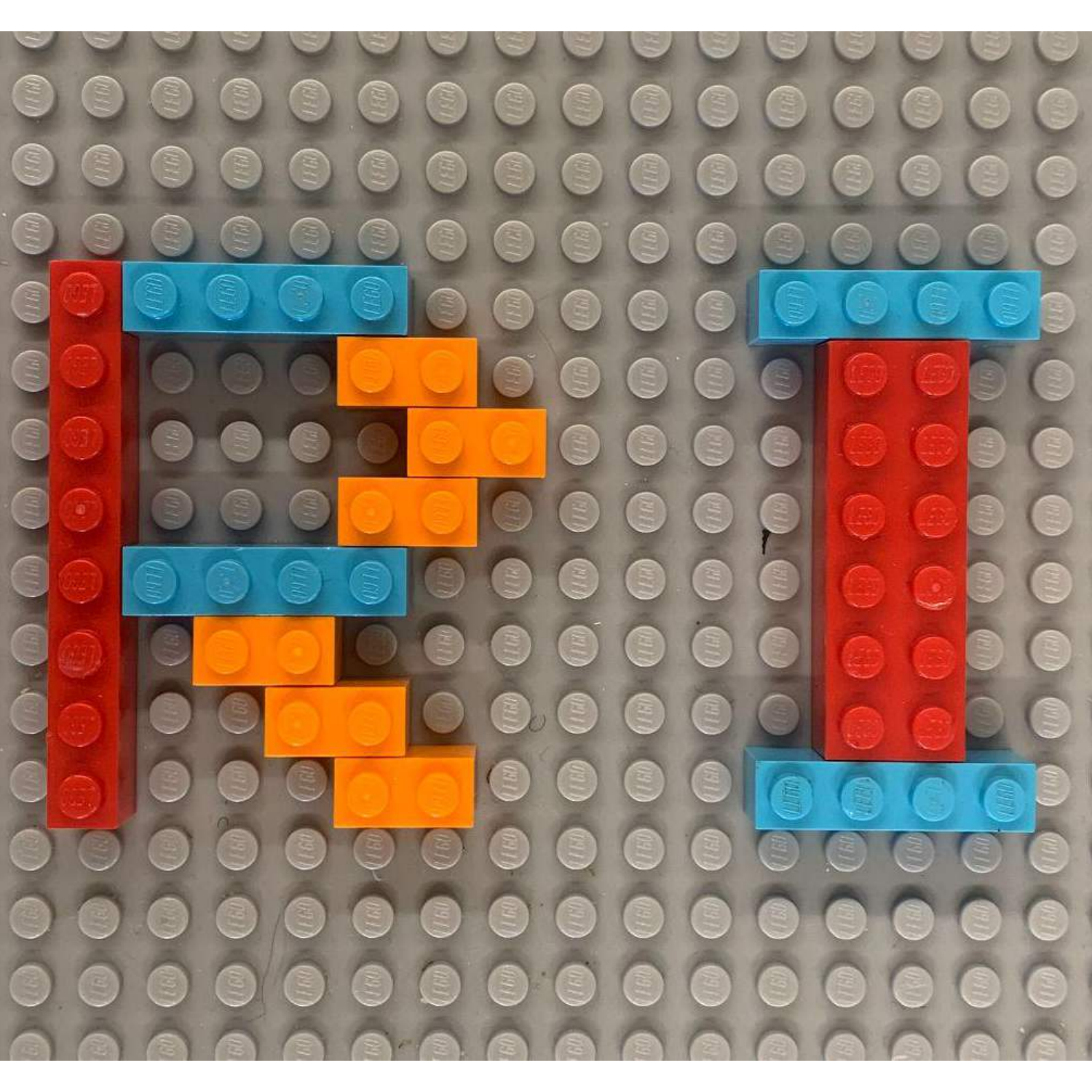}\label{fig:ri}}\hfill
\subfigure[A chair.]{\includegraphics[width=0.16\linewidth]{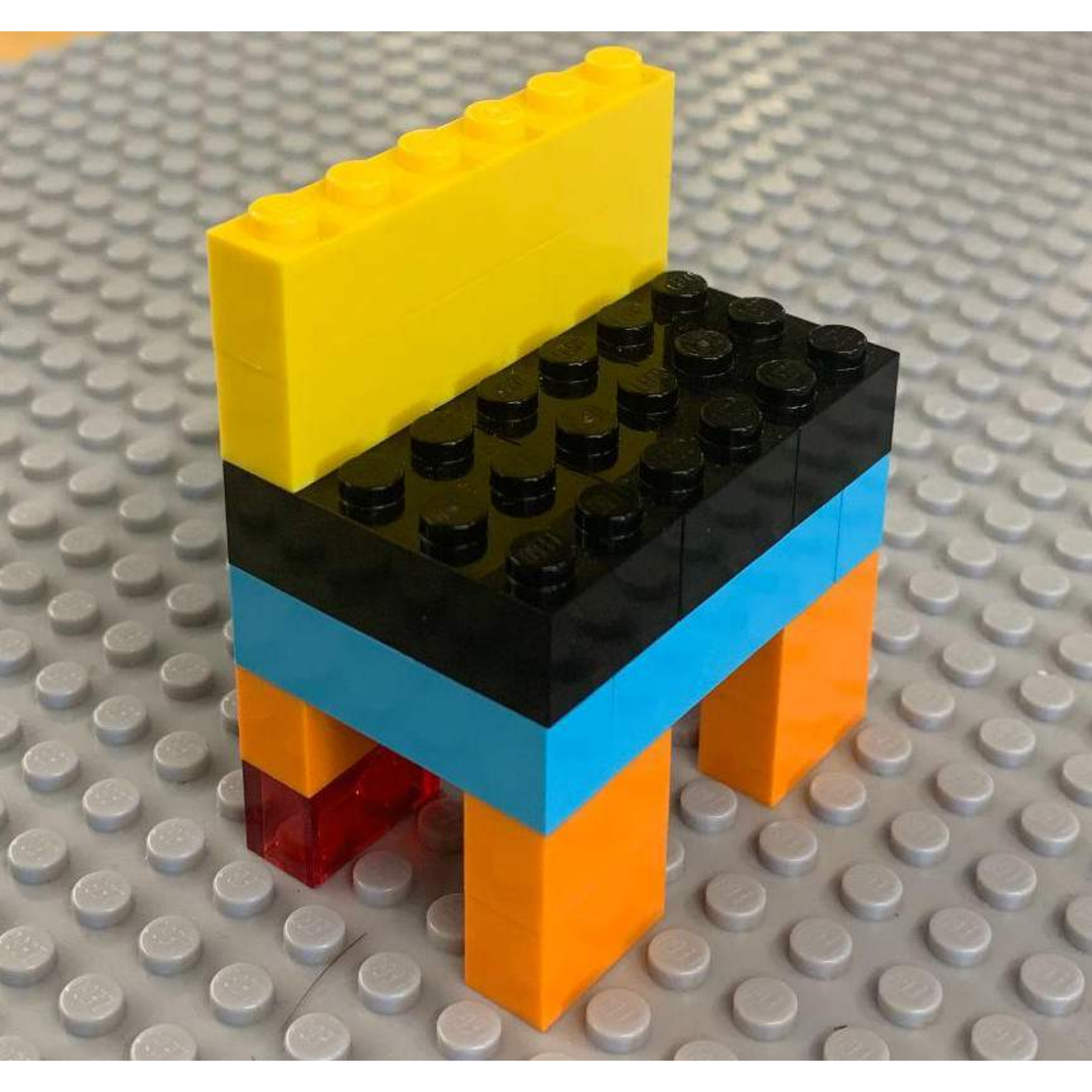}\label{fig:chair}}\hfill
\subfigure[A heart.]{\includegraphics[width=0.16\linewidth]{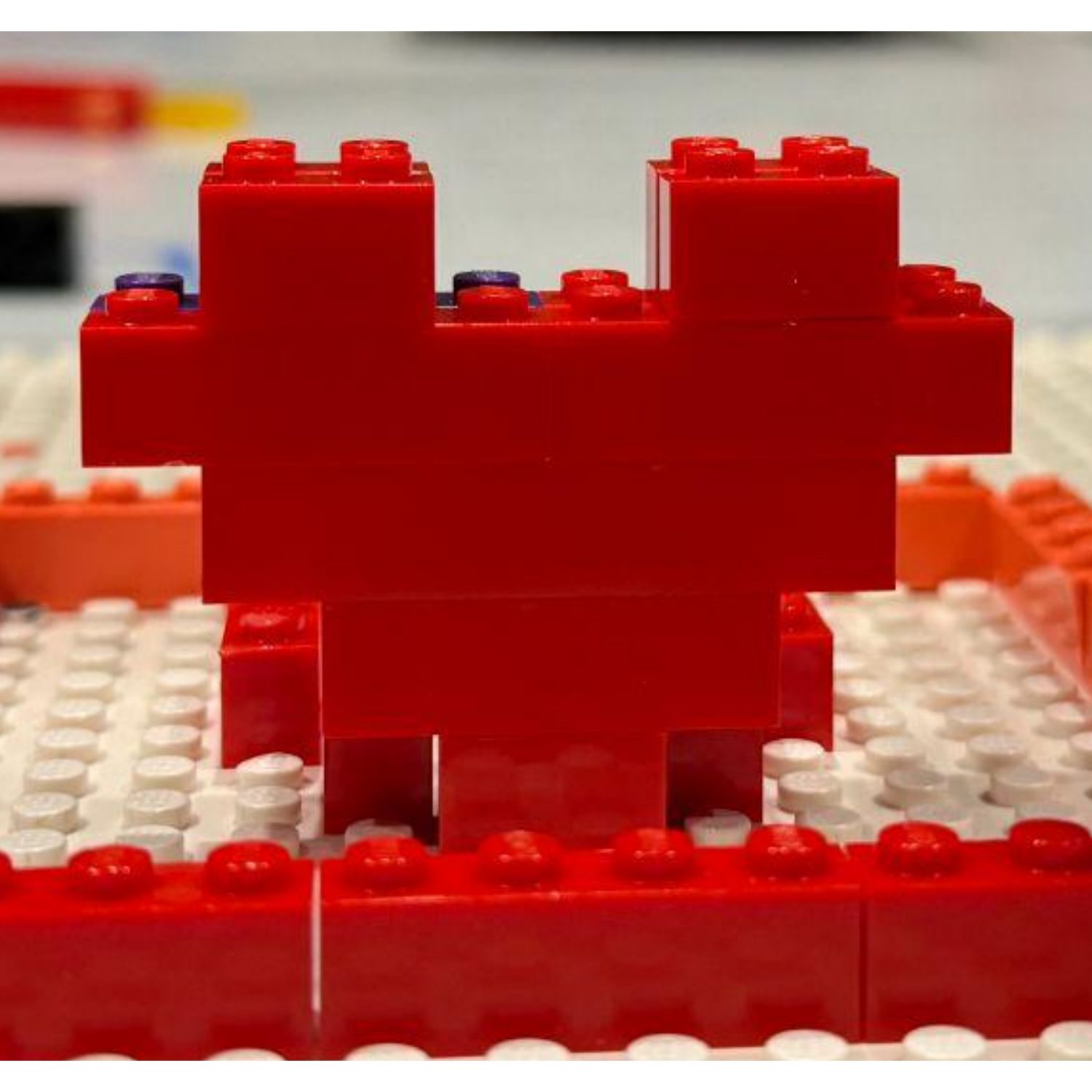}\label{fig:heart}}\hfill
\subfigure[A bridge.]{\includegraphics[width=0.16\linewidth]{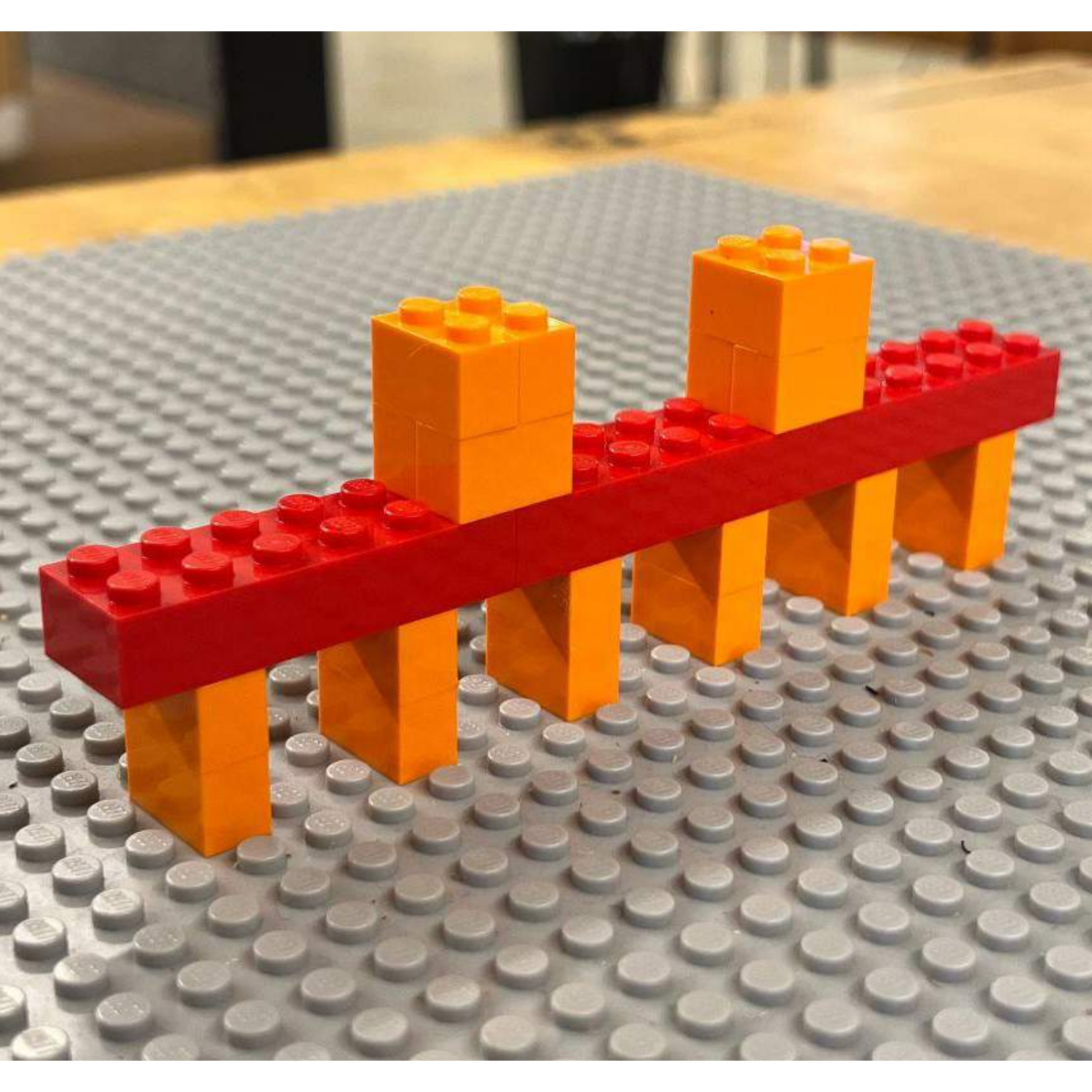}\label{fig:bridge}}\hfill
\subfigure[A temple.]{\includegraphics[width=0.16\linewidth]{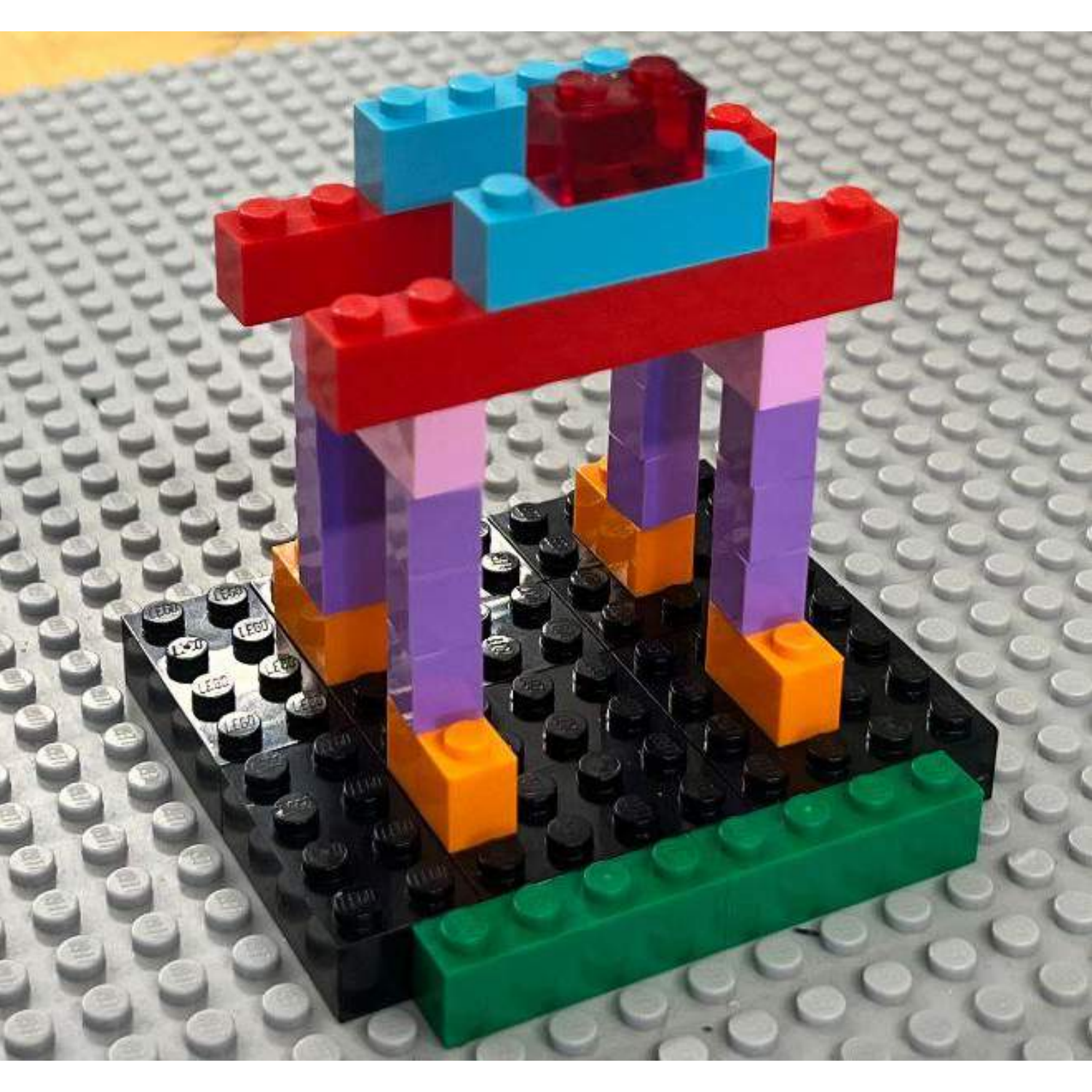}\label{fig:temple}}\\
    \caption{Top: Lego manipulation requirements. Bottom: Examples of 2D and 3D Lego prototypes.} \label{fig:prototypes}
\end{figure*}

Lego has been widely used in education since it allows children to freely create novel objects \cite{lego_spike_prime,doi:10.5772/58249}.
It is also a well-known platform for prototyping and constructing proofs-of-concept \cite{ZHOU2020103282}.
There are a wide variety of Lego bricks with different shapes and colors, which allows creative customization.
Each brick has standardized knobs that can fit the bottom of other bricks.
The bricks can be assembled by stacking the bottom of a brick onto the knobs of another brick as shown in \cref{fig:assemble_success}.
The structure can be disassembled by breaking the connection and pulling the top brick off as shown in \cref{fig:disassemble_success}.
\Cref{fig:ai,fig:ri,fig:chair,fig:heart,fig:bridge,fig:temple} illustrate examples of Lego prototypes.

Recently, Lego construction has been widely studied \cite{Kim2014SurveyOA}.
Existing works \cite{10.1145/2816795.2818091, doi:10.1177/09544054211053616,10.1111:cgf.13603,thompson2020Lego,10.1007/978-3-031-19815-1_6,wang2022translating,LegoBuilder,ChungH2021neurips,KimJ2020arxiv,lennon2021image2lego,ahn2022sequential,8419684} focus on finding a brick layout to build the target prototype. 
However, they do not consider automatic physical assembly and disassembly. 
Recent works \cite{popov2017dataefficient,10.1109/ICRA.2019.8793659,8674203}
directly learn a robot control policy for stacking Lego bricks, but it is difficult to generalize in the real world with different task specifications due to the problem complexity.
\cite{9341428,7759340,9812161} demonstrate assemblies using customized brick toys.
Despite the successful Lego manipulation, \cite{9812161,8593852} requires specially designed hardware with extra actuators.

Lego manipulation is challenging for several reasons.
First, Lego assembly requires accurate brick alignment.
\Cref{fig:assemble_fail} and \cref{fig:assemble_success} illustrate the alignment constraint. 
The connections between the top knobs and the bottom of the bricks require a tight fit.
Therefore, two bricks should be aligned well, as shown in \cref{fig:assemble_success}, to be stacked for assembly.
Slight misalignment could fail the assembly or even damage the bricks.
Second, Lego disassembly should break the structure orderly.
It is desired that the robot can disassemble one piece at a time as shown in \cref{fig:disassemble_success} instead of randomly breaking the structure as shown in \cref{fig:disassemble_fail}.
This is not trivial since it is infeasible to disassemble by directly pulling up the top brick, which would randomly drag up the bricks below it due to different tightnesses between brick connections.
It is sometimes challenging even for humans to disassemble the structure orderly due to the tight connections between bricks.
Third, Lego manipulation should be fast. 
It is desired that the robot can manipulate Lego bricks quickly in order to enable rapid prototyping.
And fourth, the manipulation should be safe.
This is important since the robot should not damage itself or the environment (\ie the workspace and Lego bricks) throughout the manipulation.

To address the challenges, this paper investigates safe and efficient Lego manipulation.
In particular, this paper leverages hardware-software co-design to reduce the complexity of the manipulation and presents a robotic solution that is capable of both assembling and disassembling standard Lego bricks. 
The manipulation capability enables fast robotic Lego prototyping.
Our contributions are as follows:
\begin{enumerate}
    \item To address the assembly alignment and 1-piece disassembly requirements, an end-of-arm tool (EOAT) is designed, which significantly reduces the manipulation complexity and allows industrial robots to easily manipulate Lego bricks.
    \item To enable safe and rapid Lego manipulation, a safe learning framework using evolution strategy is adopted to optimize the robot motion for Lego manipulation. 
    \item To demonstrate the system performance, we conduct experiments to validate the EOAT and robot learning performance in real Lego assembly and disassembly.
    To illustrate the system's potential, we enable sustainable automatic Lego prototyping by deploying the system to a FANUC LR-mate 200id/7L robot and a Yaskawa GP4 robot.
\end{enumerate}

The rest of this paper is organized as follows: 
\cref{sec:relatedworks} discusses relevant works on Lego manipulation.
\Cref{sec:EOAT} introduces the EOAT design and the mechanism for Lego manipulation.
\Cref{sec:learning} introduces the safe learning framework.
\Cref{sec:result} demonstrates the experiment results and shows sustainable robotic prototyping on several Lego prototypes.
In the end, \cref{sec:limitations} discusses the limitations and future works, and \cref{sec:conclusion} concludes the paper.

\section{RELATED WORKS}
\label{sec:relatedworks}

\begin{figure*}
\centering
\subfigure[EOAT design.]{\includegraphics[width=0.19\linewidth]{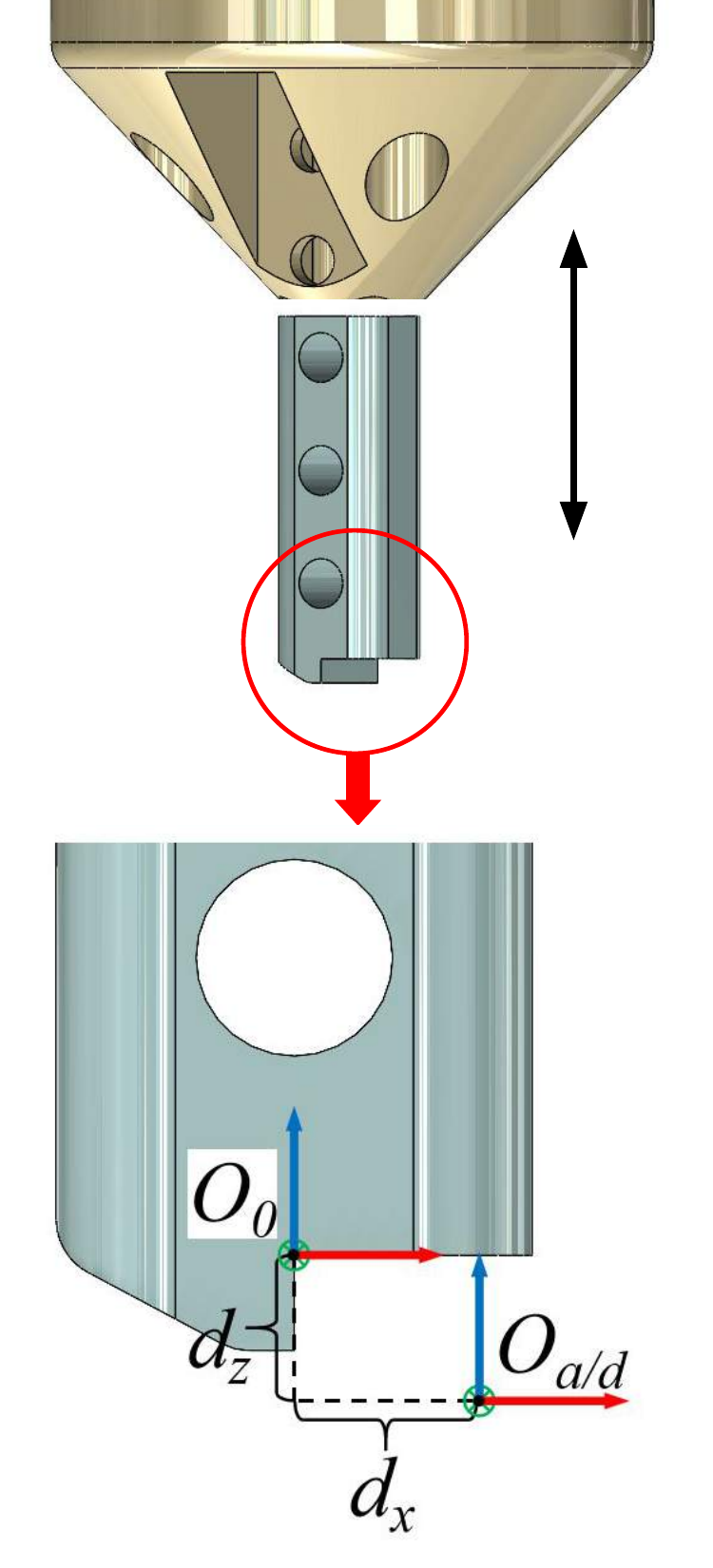}\label{fig:tool_design}}\hfill
\subfigure[Assembly and disassembly mechanism.]{\includegraphics[width=0.73\linewidth]{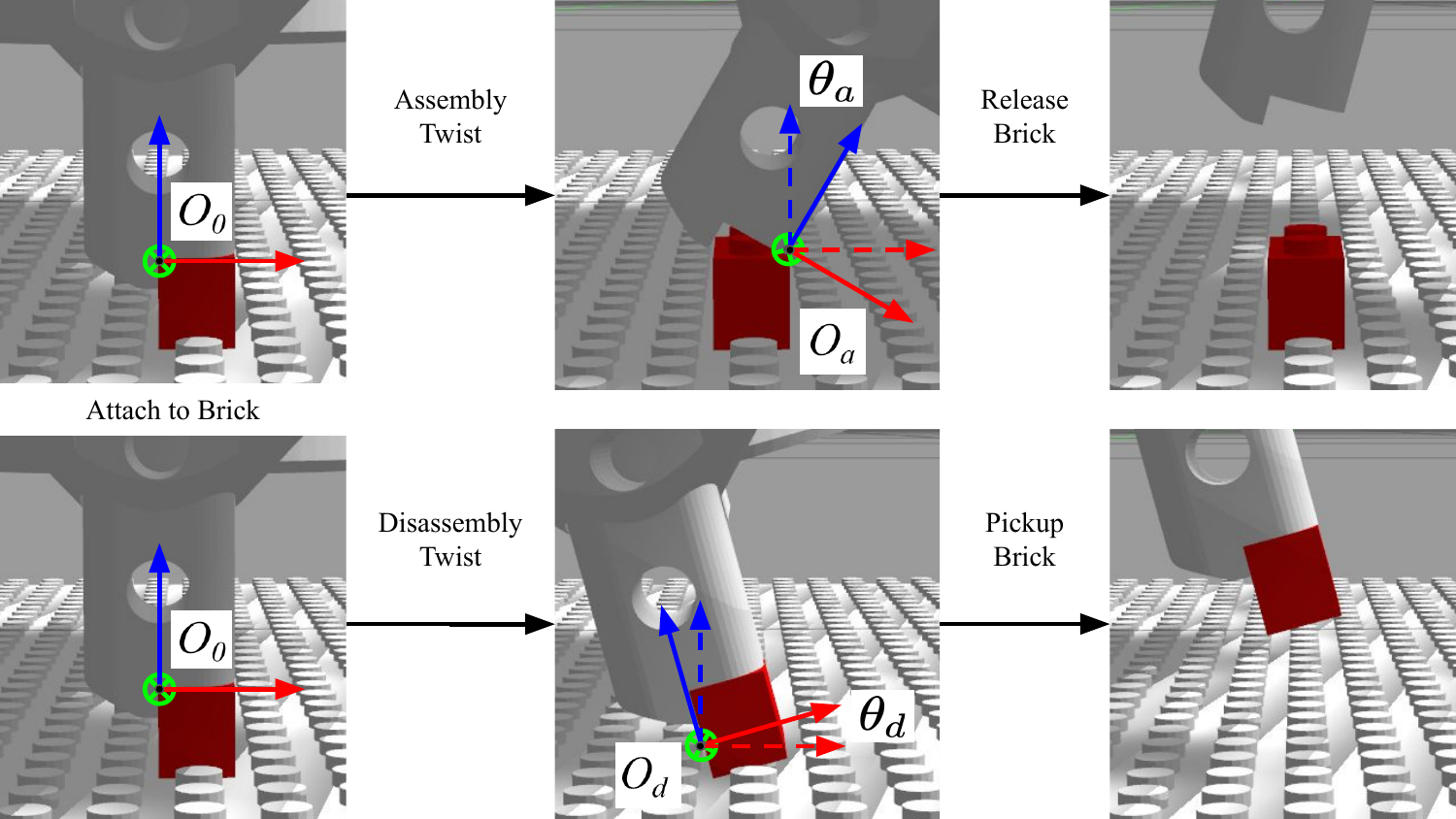}\label{fig:manipulation_mechanism}}
    \caption{EOAT design for Lego manipulation.} \label{fig:manipulation}
\end{figure*}

Existing works study Lego assembly as an insertion task, which can be considered a grasp-and-insert problem.
In the grasping stage, the robot gripper grabs the Lego brick and stacks it to the target knobs in the insertion stage.
However, it is challenging to estimate the brick pose in hand, making it difficult to align accurately for assembly. 
\cite{9341428,7759340,9812161} design specialized grippers with extra actuators for Lego grasping, which makes it easier to estimate the pose of the in-hand brick.
As a result, the assembly can be reduced to an insertion problem.
Reinforcement learning (RL) \cite{popov2017dataefficient,10.1109/ICRA.2019.8793659} can be used to end-to-end learn a control policy for the grasp-and-insert task.
However, the learning process (\ie trial-and-error exploration) could be dangerous.
Therefore, existing RL approaches train the robot in simulation \cite{stablebaselines3,zhao2023guard}, making it difficult to generalize to the real world due to the sim-to-real gap (\eg friction in brick connections).
\cite{8674203} learns the robot motion from human demonstrations.
However, it is difficult to generalize to different tasks with different settings due to the problem complexity.
On the other hand, there are few existing works addressing Lego disassembly.
\cite{8593852} designs a customized gripper with extra actuators for Lego manipulation, which is difficult to fabricate and generalize to different platforms.

\section{END-OF-ARM TOOL DESIGN}
\label{sec:EOAT}

To address both Lego assembly and disassembly, this paper formulates the manipulation as an insert-and-twist problem.
Inspired by the Lego separator \footnote{\url{https://www.lego.com/en-us/product/brick-separator-630}}, an EOAT is designed as shown in \cref{fig:tool_design}, which significantly reduces the manipulation complexity.
In particular, we use the designed EOAT to remove the in-hand pose uncertainty in grasping by inserting the Lego brick into the EOAT.
And the physical manipulation is achieved by twisting.
\Cref{fig:manipulation_mechanism} illustrates the mechanism for assembling and disassembling Lego bricks.
The EOAT first attaches to the brick by inserting the knobs of the target brick into the tool.
Due to the tight fit, the tool frame $O_0$, as shown in the bottom of \cref{fig:tool_design}, is attached to the brick as shown in the left of \cref{fig:manipulation_mechanism}.
The tight fit removes the uncertainty in grasping when estimating the in-hand brick pose.
To assemble the brick, the EOAT twists around the Y-axis of $O_a$ for $\theta_a$ and releases the brick as shown in the top row of \cref{fig:manipulation_mechanism}.
To disassemble the brick, the EOAT twists around the Y-axis of $O_d$ for $\theta_d$ and picks up the brick as shown in the bottom row of \cref{fig:manipulation_mechanism}.
Note that $O_a$ and $O_d$ are tunable by adjusting the offsets $d_x$ and $d_z$ as shown in \cref{fig:tool_design}.


\Cref{fig:tool_design} shows the overall EOAT design, which consists of two detachable components, 1) a base adaptor for mounting, and 2) a tool stick.
The modularized design has several advantages.
First, the length of the EOAT is adjustable. 
This offers the tool an extra degree of freedom to be customized and adapt to different environment settings (\eg manipulating in a sparse vs crowded setup).
Second, this modularized design allows easy and quick replacement of the tool.
During Lego manipulation, the tool stick will be worn off slowly due to contact friction.
This design allows the users to replace the worn parts without re-fabricating the entire EOAT.
Third, the design enables quick adaptation to different robots.
The modularized design allows the users to adjust the base adaptor when transitioning to different robot platforms without re-fabricating the entire EOAT.
Fourth, the modularized design makes the EOAT extensible.
External sensors (\eg wrist camera) can be integrated onboard to the base adaptor via the side slots \cite{liu2023robotic, liu2023simulation}.
And lastly, the EOAT is capable of Lego manipulation without extra actuators. The design allows easy fabrication and makes it easily transferable to different platforms.

With the EOAT design, the complexity of Lego manipulation is greatly reduced.
The robot manipulation can be explicitly defined using $d_x, d_z$, and $\theta$.
However, $d_x, d_z,$ and $\theta$ are tunable, which could influence the manipulation performance significantly.
Therefore, we propose to learn the parameters to optimize (\ie safe and fast) the assembly and disassembly performance.


\begin{algorithm}[t]
    \caption{Safe Robot Learning for Lego Manipulation with CMAES}
    \label{alg:cmaes}
	\begin{algorithmic}[1]
	    \State Initialize $\alpha, \beta, \gamma, \eta$. 
            \State Initialize $T_0, \theta_0, d_{x0}, d_{z0}$ based on prior knowledge.
            \State Initialize optimizer CMAES$(T_0, \theta_0, d_{x0}, d_{z0}, \sigma)$.
    
            \For {$epoch=1,2,\ldots, M$}
                \State Reset $solution\_set = \emptyset$.
                \State Obtain proposed seeds $S$ from CMAES.
                \For {$s=\{T', \theta';d_x';d_z'\}$ in $S$}
                    \State Execute the assembly/disassembly tasks.
                    \State Calculate costs in \cref{eq:cost}.
                    \State Append costs to $solution\_set$.
    	    \EndFor
    	    \State CMAES updates $T, \theta, d_x, d_z$ based on $solution\_set$.
		\EndFor
            \State Return $T,\theta, d_x, d_z$.
	\end{algorithmic} 
\end{algorithm}

\section{ROBOT LEARNING FOR LEGO MANIPULATION}
\label{sec:learning}

The designed EOAT allows the robot to manipulate (\ie assemble and disassemble) Lego bricks as shown in \cref{fig:manipulation} by inserting and twisting.
However, it remains challenging to tune the parameters $d_x, d_z,$ and $\theta$ for fast and safe manipulation.
To optimize the manipulation performance, we propose to learn these parameters \cite{zhao2024autonomous}.
The learning problem can be formulated as 
\begin{equation}
\label{eq:obj}
    \begin{split}
        \min_{T, d_x,d_z,\theta} &L(\theta, d_x, d_z, T, U),\\ 
        ~\st~ & U=[u_1;u_2;\dots;u_T]\in \mathbf{R}^{n\times T},\\
        ~& u_t \in [u_{min}, u_{max}],
        ~\theta \in [\theta_{min}, \theta_{max}],\\
        ~&d_x \in [d_{xmin}, d_{xmax}],
         ~d_z\in [d_{zmin}, d_{zmax}],
    \end{split}
\end{equation}
where $T$ is the task horizon and $n$ is the robot degree of freedom. $U$ is the control sequence generated by the robot controller to execute the manipulation.
$u_{min}$ and $u_{max}$ are the control limit of the controller.
$[\theta_{min}, \theta_{max}], ~[d_{xmin},d_{xmax}],$ and $[d_{zmin},d_{zmax}]$ are the bounds on the twisting angle and axis offsets, which can be obtained from prior task knowledge (\eg Lego dimensions, robot reachability).
$L(\cdot)$ is a cost function, which is defined as
\begin{equation}
\label{eq:cost}
    \begin{split}
        L(\theta,d_x,d_z,T,U)=\begin{cases}
			&\alpha T+\beta \theta + \gamma F + \frac{\eta}{T}\sum^T_1|u_t|, \\ 
            & \text{if manipulation succeeded.}\\
            &\infty,\\ 
            & \text{otherwise.}
		 \end{cases}
    \end{split}
\end{equation}
where $F$ is the force feedback.
$\alpha, \beta, \gamma,\eta$ are tunable weights to adjust the penalty on each term during the learning process.
Note that different robot controllers would generate different $U$ to execute the manipulation task.
Therefore, \cref{eq:cost} considers the controller output $U$.
The objective is to have the robot finish the task in a shorter time (\ie $\alpha T$), with a shorter travel distance (\ie $\beta \theta$), with less impact on the Lego structure (\ie $\gamma F$), and with smoother robot motion (\ie $\frac{\eta}{T}\sum^T_1|u_t|$).
Since it is difficult to obtain the gradient of \cref{eq:cost}, we solve \cref{eq:obj} using evolution strategy, in particular, covariance matrix adaptation evolution strategy (CMAES) \cite{hansen2016cma,nomura2021warm}.
The proposed learning solution to \cref{eq:obj} is summarized in \cref{alg:cmaes}.
Based on the prior task knowledge, we initialize the optimizer with initial guesses on $T_0,\theta_0, d_{x0}, d_{z0}$ and standard deviation $\sigma$ on lines 2-3.
For each epoch, we obtain proposed solutions on line 6 and execute the Lego assembly or disassembly on line 8. Based on the execution results, we obtain the costs according to \cref{eq:cost} on lines 9.
The optimizer updates the parameters' mean and covariance matrices and adjusts its searching step based on the observed cost values on line 12.

\begin{figure}
\subfigure[]{\includegraphics[width=0.49\linewidth]{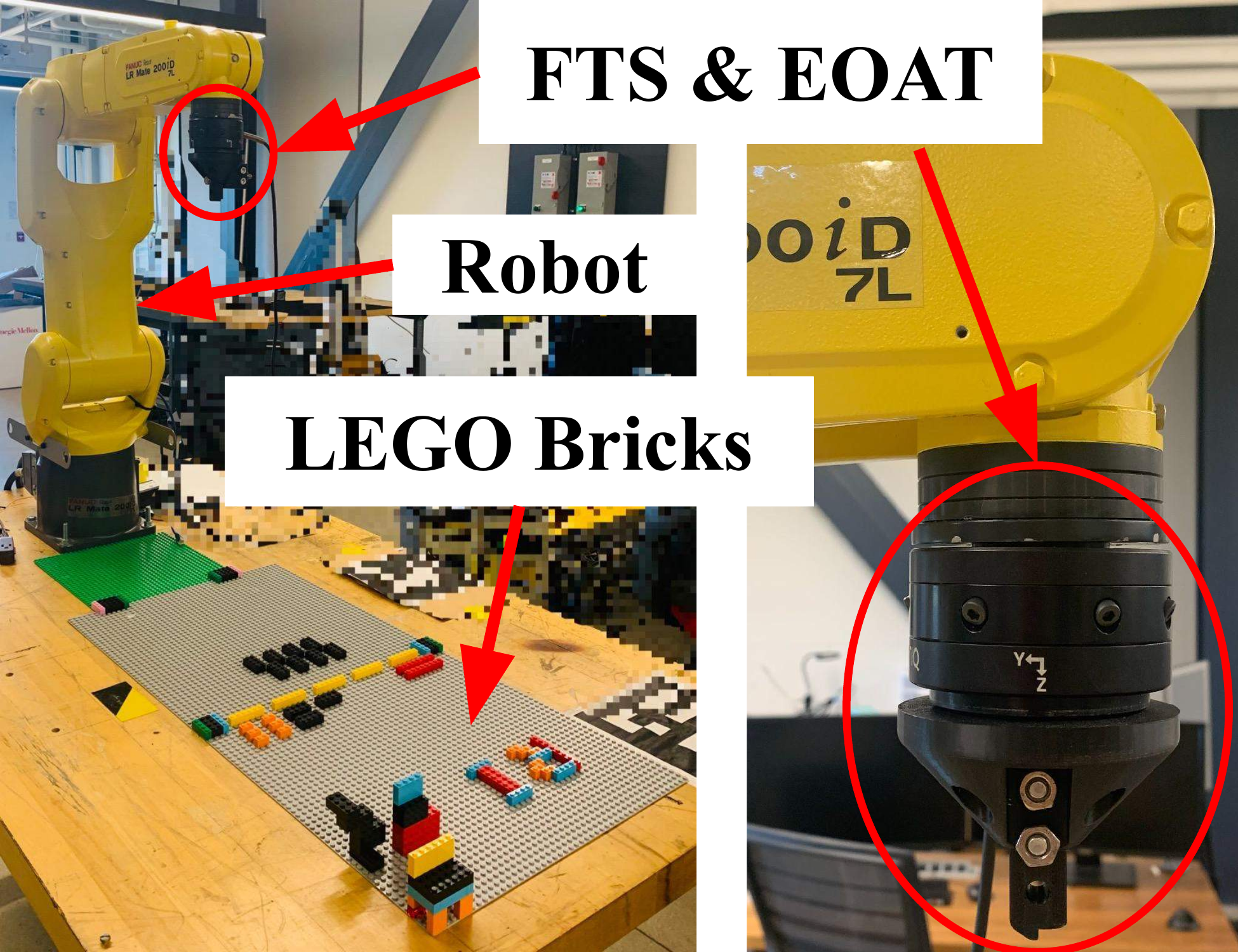}\label{fig:env}}\hfill
\subfigure[]{\includegraphics[width=0.49\linewidth]{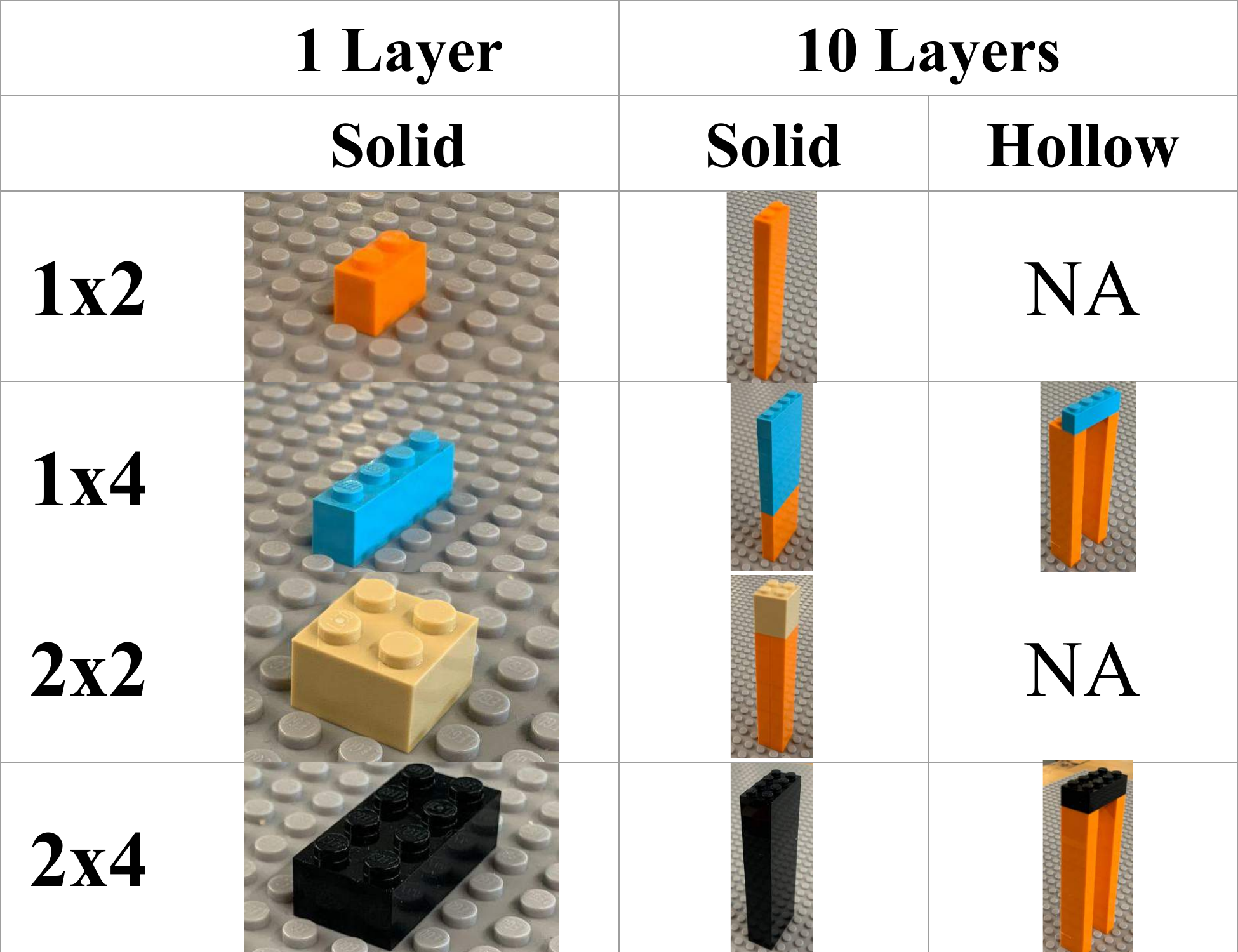}\label{fig:manipulation_structures}}
    \caption{Left: Experiment environment setup. Right: Lego structures for testing manipulation performance.} \label{fig:exp_env}
\end{figure}

\begin{table*}
\footnotesize
\centering
\begin{tabular}{c  c  c  c  c  c c c c c} 
\hline
Brick & Height & Support & Controller & Success Rate & Brick & Height & Support & Controller & Success Rate\\   
\hline
\multirow{8}{*}{$1\times 2$} & \multirow{4}{*}{1} & \multirow{4}{*}{Solid} & \multirow{2}{*}{Joint JPC} &  (\textbf{100\%}  /  96\%) & \multirow{8}{*}{$2\times 2$} & \multirow{4}{*}{1} & \multirow{4}{*}{Solid} & \multirow{2}{*}{Joint JPC}   &  (\textbf{100\%}  /  94.8$\%$)\\ 
 &  &  &  & [\textbf{100\%}/\textbf{100\%}] & &  &  &  &   [\textbf{100\%}  /  \textbf{100\%}]\\ 
 &  &  & \multirow{2}{*}{Cartesian JPC}  & (\textbf{100\%}  /  86.8\%) & &  &  & \multirow{2}{*}{Cartesian JPC} & (\textbf{100\%} /  98\%)\\
  &  &  &  &   [\textbf{100$\%$}  /  \textbf{100$\%$}] & &  &  &  &    [\textbf{100\%}  /  \textbf{100\%}]\\ 
  \cline{2-5}
  \cline{7-10}
 & \multirow{4}{*}{10} & \multirow{4}{*}{Solid} & \multirow{2}{*}{Joint JPC} & (\textbf{100\%}  /  89.2\%) & & \multirow{4}{*}{10} & \multirow{4}{*}{Solid} & \multirow{2}{*}{Joint JPC} & (\textbf{100\%} /  93.2\%)\\ 
 &  &  &  & [\textbf{100\%}  /  \textbf{100\%}] &  &  &  &   &  [\textbf{100\%}  /  \textbf{100\%}]\\ 
 &  &  & \multirow{2}{*}{Cartesian JPC} &    (\textbf{100\%}  /  85.2$\%$) & &  &  & \multirow{2}{*}{Cartesian JPC} & (\textbf{100\%}  /  94\%)\\ 
  &  &  &  &  [\textbf{100\%}  /  \textbf{100\%}] & &  &  &  & [\textbf{100\%}  /  \textbf{100\%}]\\ 
\hline
\multirow{12}{*}{$1\times 4$} & \multirow{4}{*}{1} & \multirow{4}{*}{Solid} & \multirow{2}{*}{Joint JPC} &  (\textbf{100\%}  /  92\%) & \multirow{12}{*}{$2\times 4$} & \multirow{4}{*}{1} & \multirow{4}{*}{Solid} & \multirow{2}{*}{Joint JPC}   & (\textbf{100\%}  /  96\%)\\ 
 &  &  &  &   [\textbf{100\%}  /  \textbf{100\%}] & &  &  &  & [\textbf{100\%}  /  \textbf{100\%}]\\ 
 &  &  & \multirow{2}{*}{Cartesian JPC} &  (\textbf{100\%}  /  93.2\%) & &  &  & \multirow{2}{*}{Cartesian JPC} & (\textbf{100\%}  /  92.4\%)\\ 
  &  &  &  &       [\textbf{100\%}  / \textbf{100\%}] & &  &  &  & [\textbf{100\%} /  \textbf{100\%}]\\ 
  \cline{2-5}
   \cline{7-10}
 & \multirow{8}{*}{10} & \multirow{4}{*}{Solid} & \multirow{2}{*}{Joint JPC} &  (\textbf{100\%}  /  89.6\%) & & \multirow{8}{*}{10} & \multirow{4}{*}{Solid} & \multirow{2}{*}{Joint JPC} & (\textbf{100\%} /  95.2\%)\\ 
 &  &  &  & [\textbf{100\%}  /  \textbf{100\%}] & &  &  &  & [\textbf{100\%}  /  \textbf{100\%}]\\ 
 &  &  & \multirow{2}{*}{Cartesian JPC} & (\textbf{100\%}  /  90.8\%) &  &  &  & \multirow{2}{*}{Cartesian JPC} & (\textbf{100\%}  /  98\%)\\ 
  &  &  &  & [\textbf{100\%}  /  \textbf{100\%}] & &  &  &  & [\textbf{100\%}  /  \textbf{100\%}]\\ 
  \cline{3-5}
  \cline{8-10}
 &  & \multirow{4}{*}{Hollow} & \multirow{2}{*}{Joint JPC} &     (\textbf{100\%}  /  88\%) & &  & \multirow{4}{*}{Hollow} & \multirow{2}{*}{Joint JPC} & (\textbf{100\%}  /  94.8\%)\\ 
  &  &  &  &    [\textbf{100\%}  /  \textbf{100\%}] &  &  &  &  &   [\textbf{100\%}  /  \textbf{100\%}]\\ 
 &  &  & \multirow{2}{*}{Cartesian JPC} & (\textbf{100\%}  /  83.6\%) & &  &  & \multirow{2}{*}{Cartesian JPC} & (\textbf{100\%} /  84.4\%)\\ 
  &  &  &  &       [\textbf{100\%} /  \textbf{100\%}] & &  &  &  & [\textbf{100\%}  /  \textbf{100\%}]\\ 
\hline
\end{tabular}
\caption{Success rate of Lego brick manipulation. $(\cdot/\cdot):$ assembling and disassembling success rate without safe learning. $[\cdot/\cdot]:$ optimized assembling and disassembling success rate with safe learning.}\label{table:EOAT_performance}
\end{table*}

Note that \cref{alg:cmaes} is executed directly on the physical robot.
This is because, to the best of our knowledge, no existing simulators can simulate the connections between bricks, and therefore, it is challenging to calculate \cref{eq:cost} without executing in real.
However, directly applying RL methods to real robots can cause safety hazards due to trial-and-error exploration.
\Cref{alg:cmaes} is safe and can be directly applied to physical robots for the following reasons.
First, due to the EOAT design,
\cref{alg:cmaes} constrains the robot learning space using prior knowledge on the task, \ie bounds on the parameters.
Thus, the learning space is reduced, and thus, safe for exploration.
Second, the idea of hardware and software co-design \cite{chen2022composable} reduces the sim-to-real gap and enables direct learning on physical robots.
Third, \cref{alg:cmaes} is controller independent. 
Therefore, we can learn the manipulation parameters on top of any robot controllers, which can be safe controllers \cite{9029720, jpc, jssa, ssa,1087247,6414600,7040372,taskagnostic,10252579,wei2023zero,chen2024real} that inherently ensure safety.
Fourth, due to the consideration of extra sensor feedback (\ie force feedback), the robot can terminate the exploration as soon as $F$ reaches abnormal values to ensure safe exploration.

\section{EXPERIMENTS}
\label{sec:result}

To demonstrate the proposed system, we fabricated the EOAT using 3D printing and deployed it to a FANUC LR-mate 200id/7L industrial robot.
\Cref{fig:env} illustrates the experiment setup.
The EOAT is mounted to a Robotiq force-torque sensor (FTS), which is then mounted to the robot end-effector.
The detailed installation and hardware demonstration are shown in the video at \url{https://youtu.be/CqrBryAGsd0?si=ICyQNfXSHD6lc6OI}.
The robot builds and breaks Lego bricks on the $48\times 48$ Lego plates placed in front of it.
Based on empirical observations, we initialize $T_0=2\si{s}$, $\theta_0=15^{\circ}$, and $d^a_{x0}=7.8\si{mm}, d^a_{z0}=0\si{mm}$ for assembly and $d^d_{x0}=0\si{mm}, d^d_{z0}=3.2\si{mm}$ for disassembly.
We have $d_{x0}^a=7.8\si{mm}$ because the length of the top lever of the tool is $7.8\si{mm}$.
$d_{z0}^d$ is initialized to $3.2\si{mm}$ since the length of the side lever of the tool in \cref{fig:tool_design} is $3.2\si{mm}$.
We perform assembly and disassembly on different bricks (\ie $1\times 2$, $1\times 4$, $2\times 2$, $2\times 4$) with different heights (\ie 1-10 layers) with different supporting structures (\ie solid and hollow), as shown in \cref{fig:manipulation_structures}, across 25 positions on the plate.
To demonstrate the generalizability of our system (\ie controller independent), we implemented two robot controllers, 1) a joint-based jerk-bounded position controller (JPC) \cite{jpc} and 2) a Cartesian-based JPC, which is a derivation of the original JPC.

\begin{figure*}
    \centering
    \input{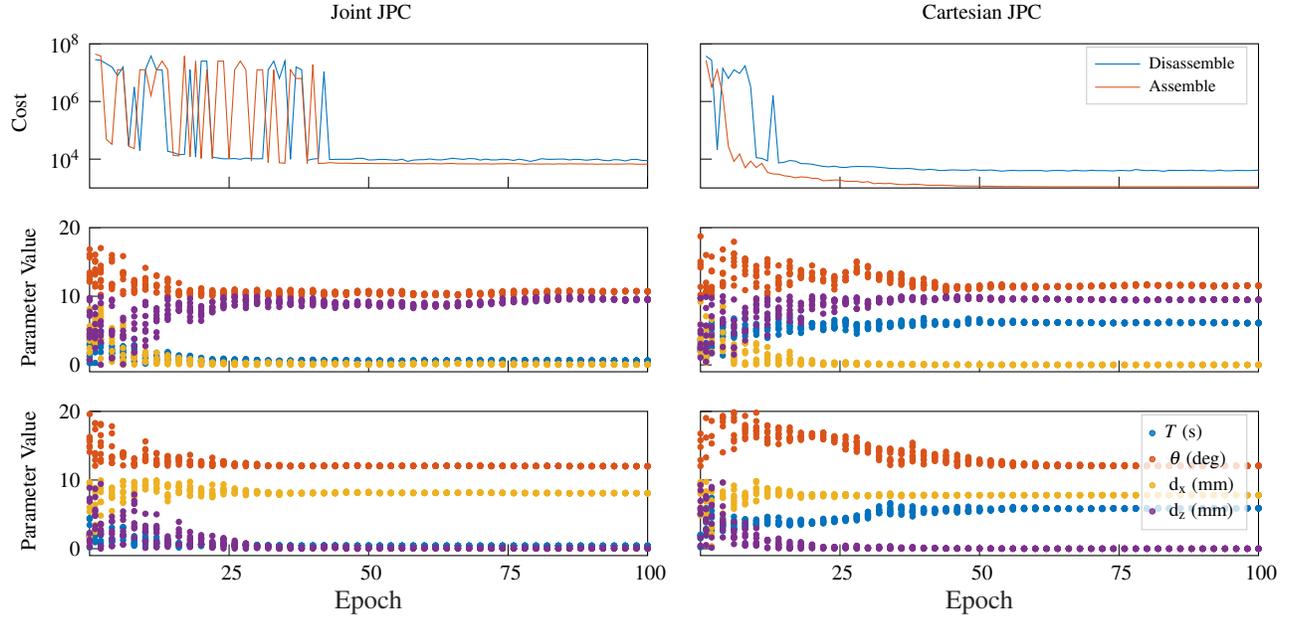}
    \caption{Safe manipulation learning costs and parameters evolution. Left: Joint JPC. Right: Cartesian JPC. Top: learning costs. Middle: disassembly parameters. Bottom: assembly parameters. Joint JPC disassembly parameters: $T=0.6$\si{s}, $\theta=11^\circ$, $d_x=0$\si{mm}, $d_z=9.6$\si{mm}. Joint JPC assembly parameters: $T=0.4$\si{s}, $\theta=12^\circ$, $d_x=8.1$\si{mm}, $d_z=0$\si{mm}. Cartesian JPC disassembly parameters: $T=6.1$\si{s}, $\theta=12^\circ$, $d_x=0$\si{mm}, $d_z=9.5$\si{mm}. Cartesian JPC assembly parameters: $T=5.9$\si{s}, $\theta=12^\circ$, $d_x=7.8$\si{mm}, $d_z=0$\si{mm}.}
    \label{fig:learning_costs}
\end{figure*}

\begin{figure}
\centering
\includegraphics[width=0.85\linewidth]{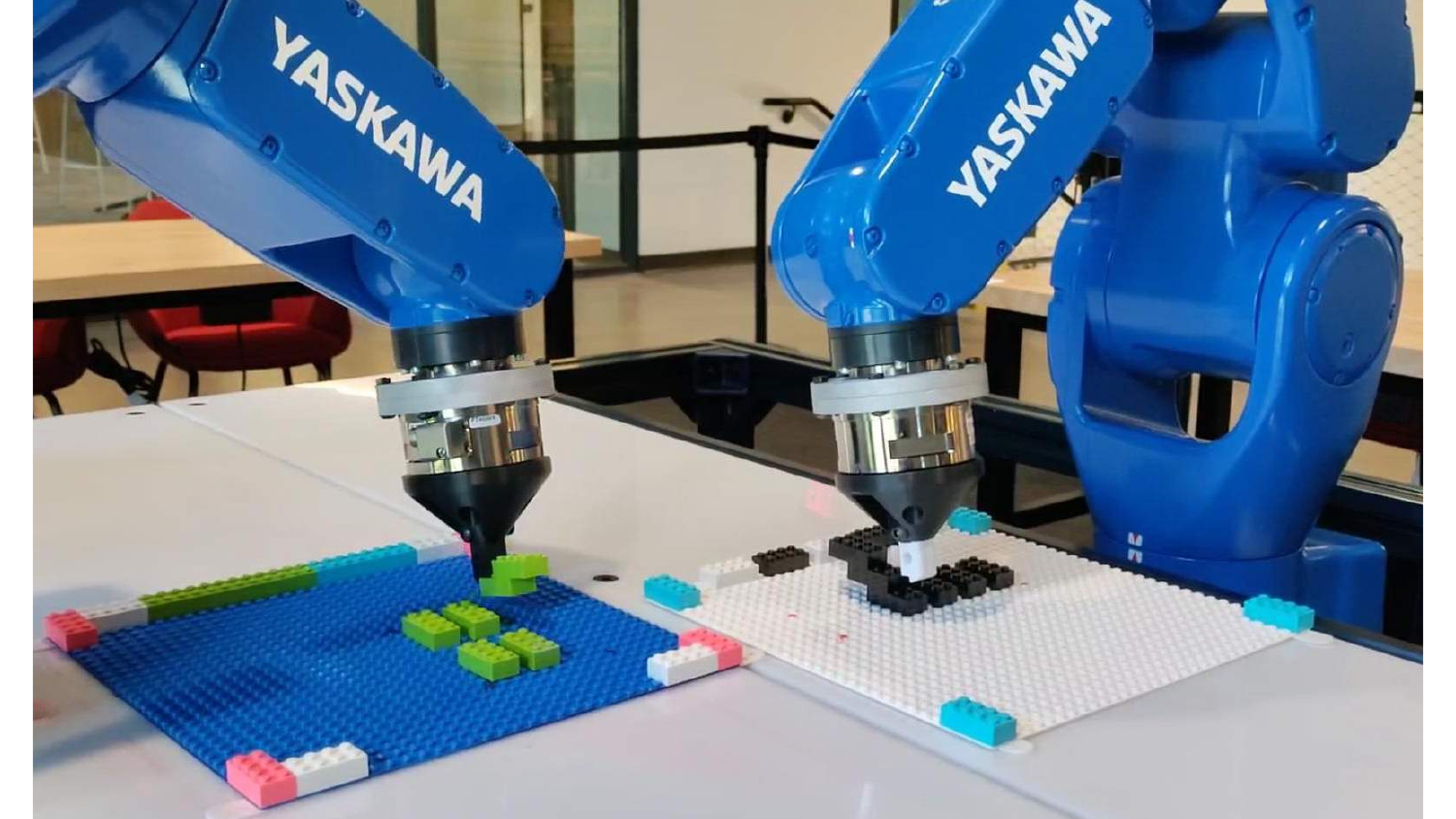}
    \caption{Transferable to different robot platforms: Yaskawa GP4 robots manipulating Lego bricks.} \label{fig:yaskawa}
\end{figure}

\Cref{table:EOAT_performance} illustrates the EOAT performance. 
The success rate of each scenario is calculated over the 25 different positions on the plate.
The robot manipulates for 10 trials at each position for each configuration.
By using the EOAT with the empirical parameters, we observe that the robot is capable of successfully manipulating Lego bricks. 
It achieves 100\% success rate in assembling and a decent success rate ($>$80\%) in disassembling Lego bricks.
It performs well in assembling due to the hardware design, which greatly reduces the manipulation complexity.

\begin{figure*}
\centering
\subfigure[]{\includegraphics[width=0.16\linewidth]{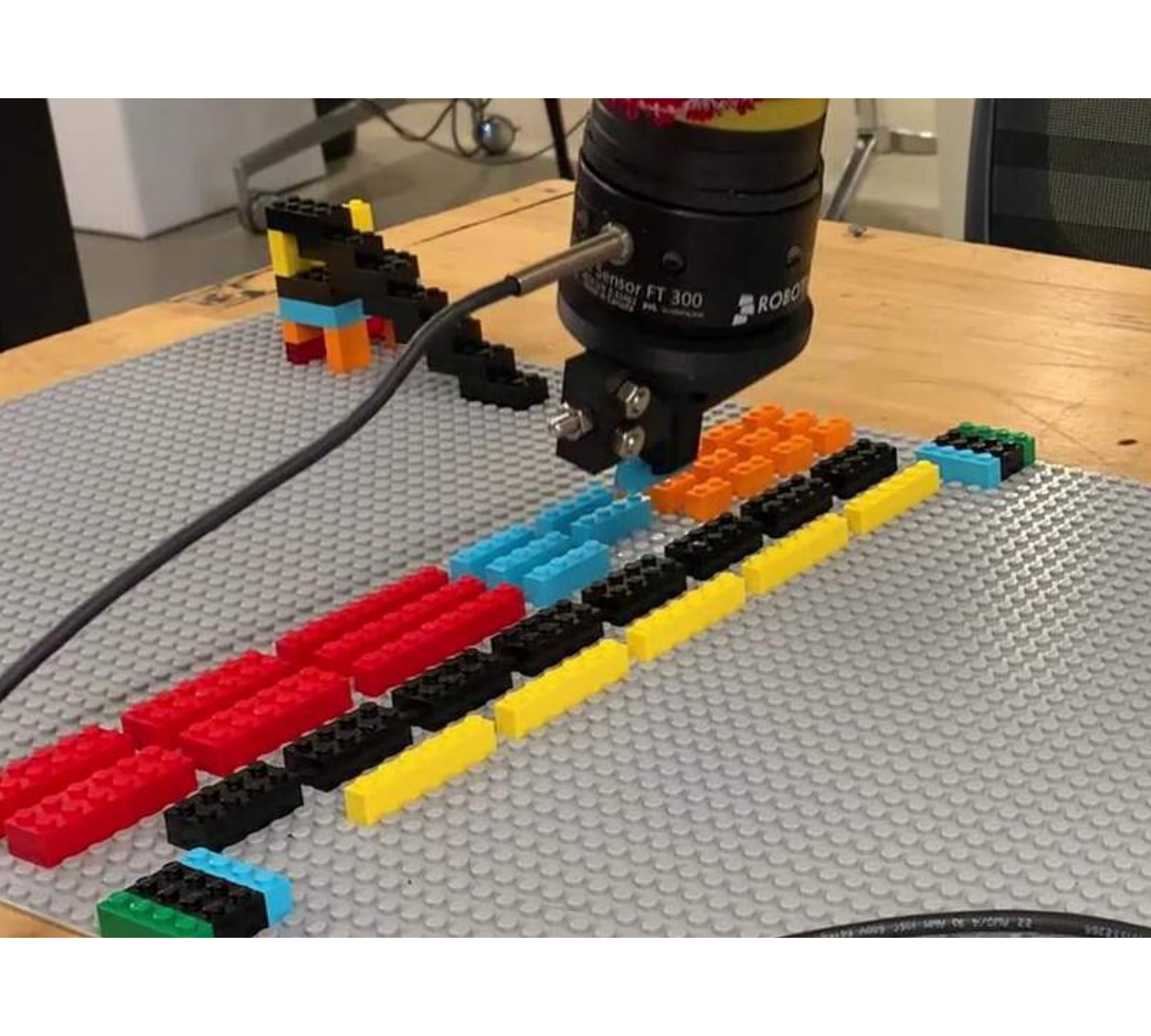}\label{fig:1}}\hfill
\subfigure[]{\includegraphics[width=0.16\linewidth]{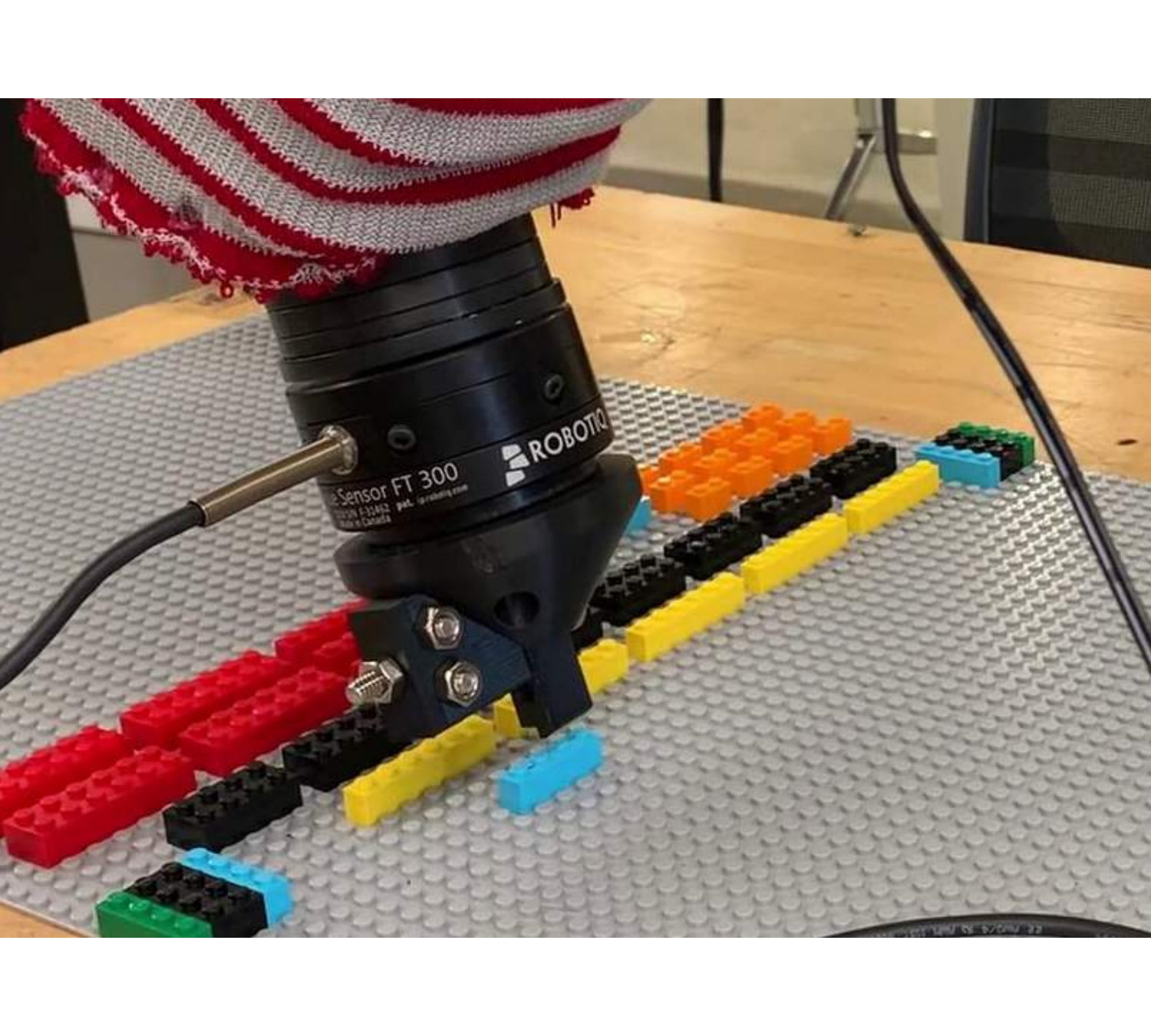}\label{fig:2}}\hfill
\subfigure[]{\includegraphics[width=0.16\linewidth]{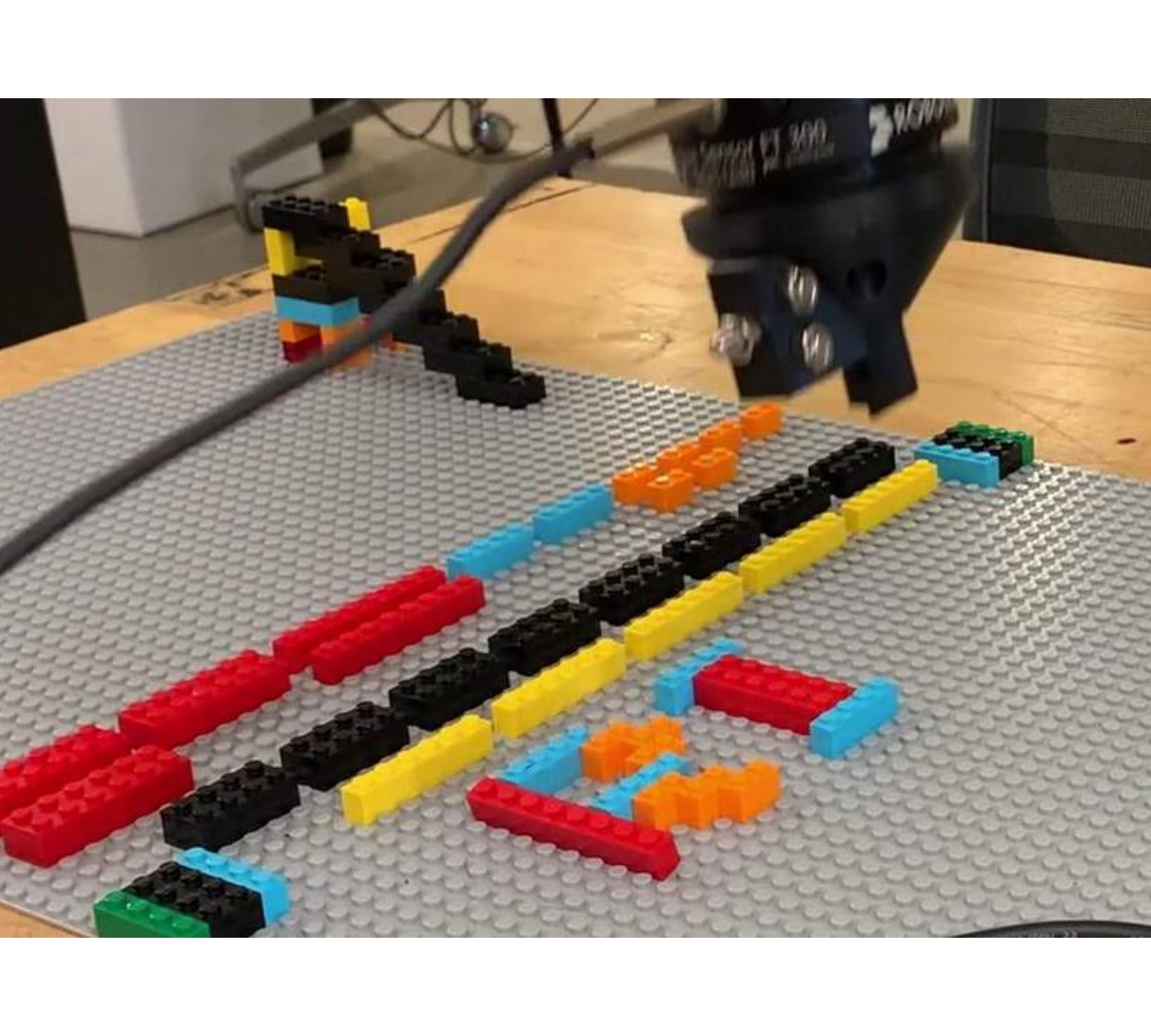}\label{fig:3}}\hfill
\subfigure[]{\includegraphics[width=0.16\linewidth]{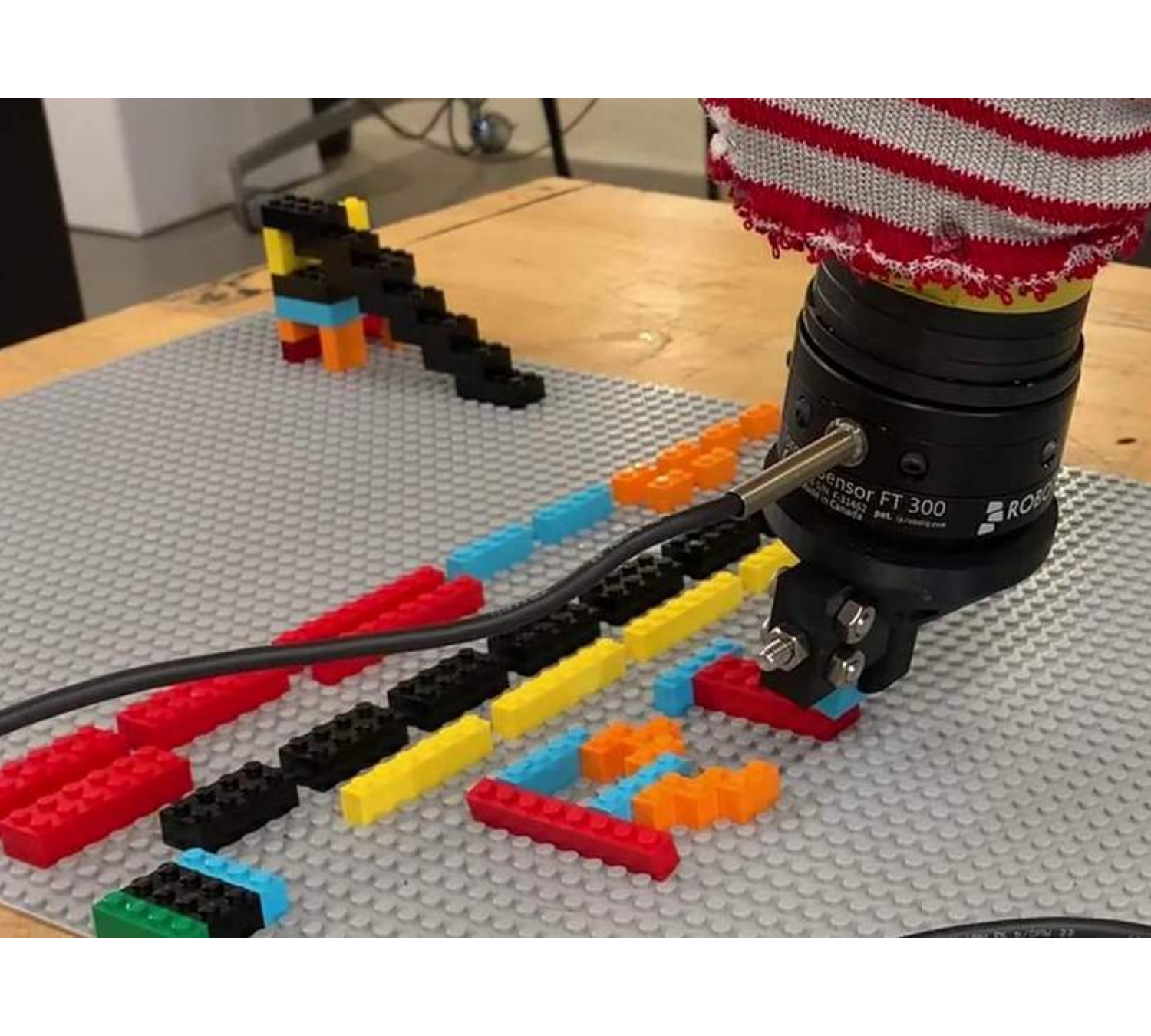}\label{fig:4}}\hfill
\subfigure[]{\includegraphics[width=0.16\linewidth]{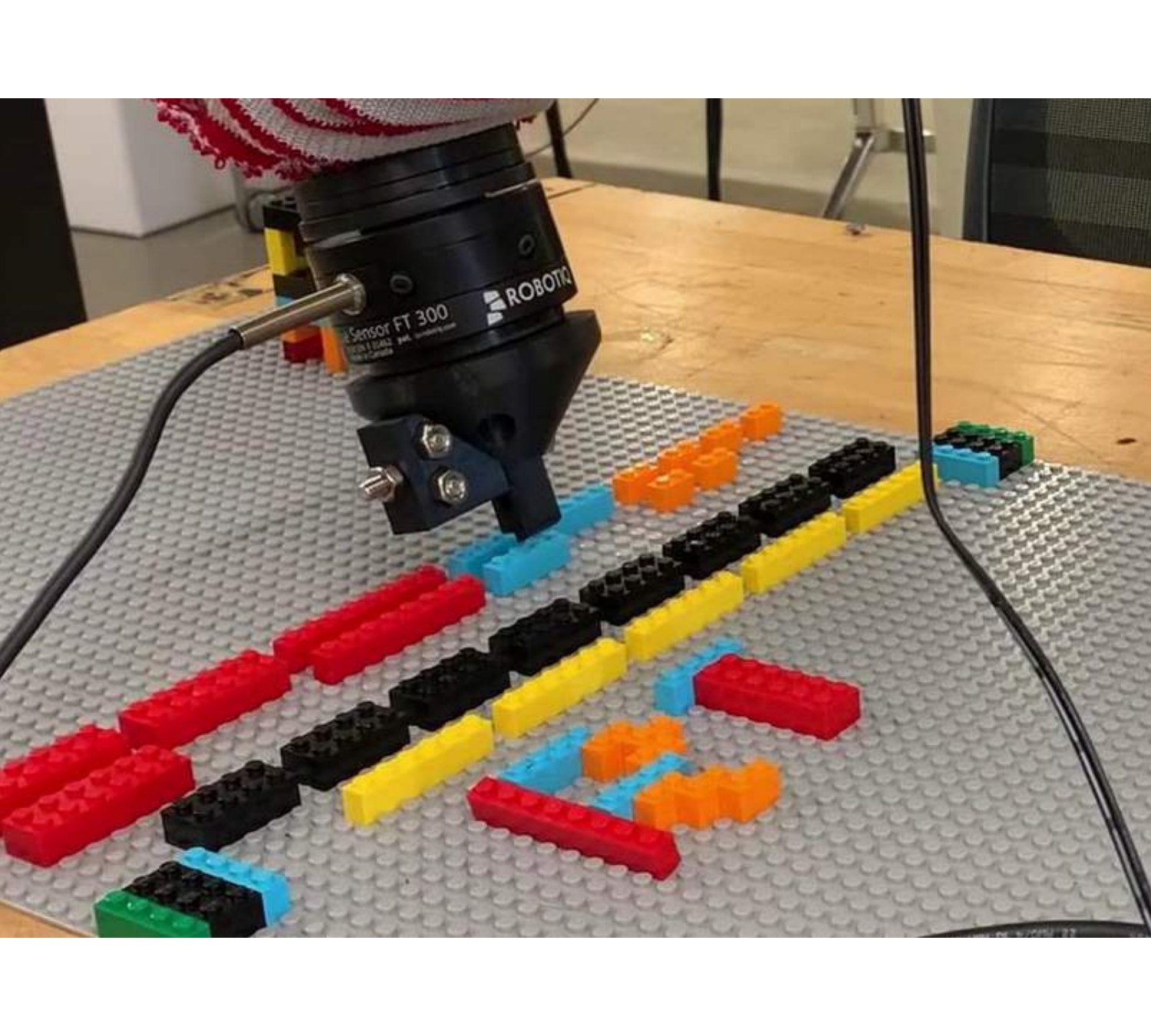}\label{fig:5}}\hfill
\subfigure[]{\includegraphics[width=0.16\linewidth]{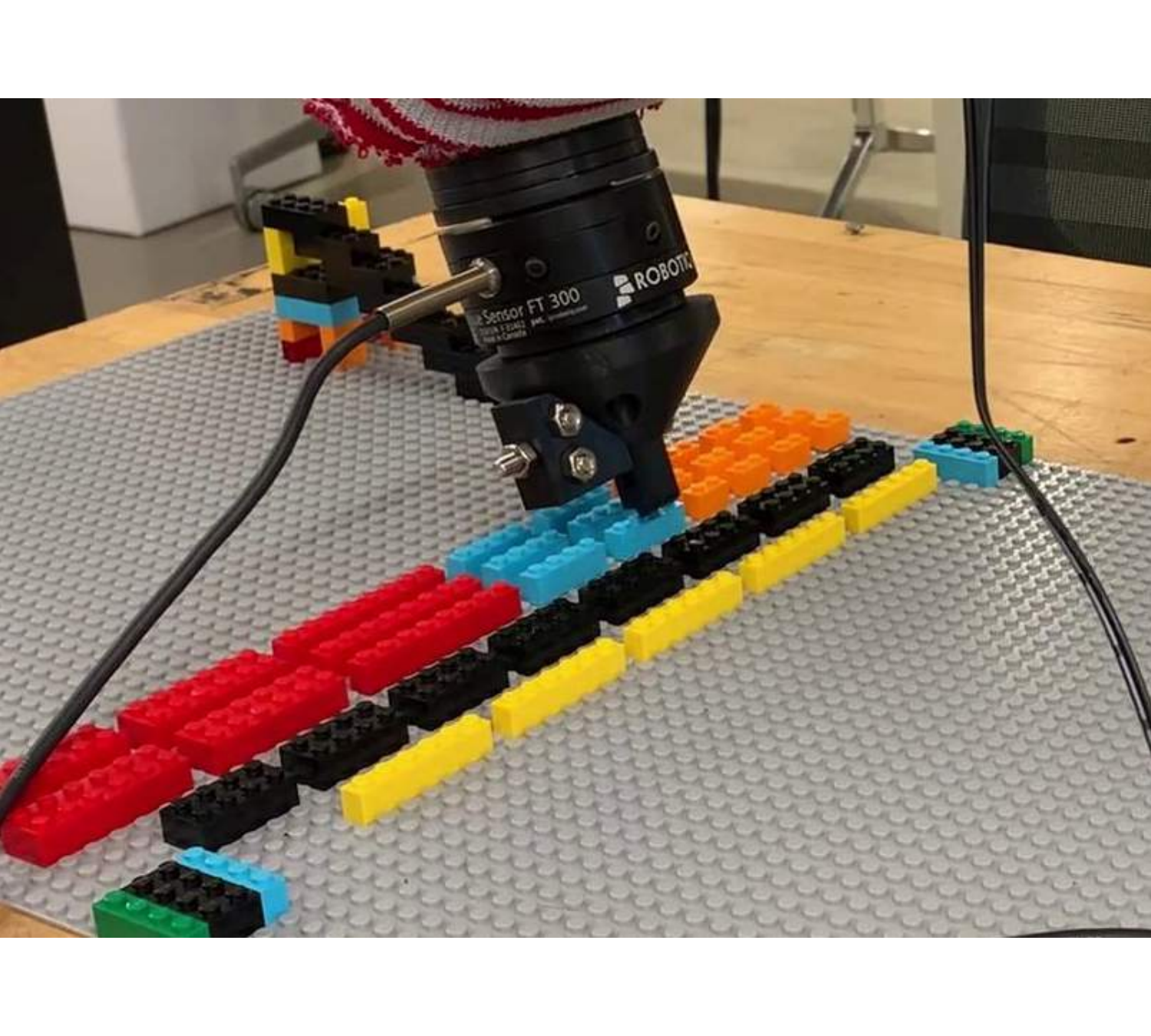}\label{fig:6}}
\\
\subfigure[]{\includegraphics[width=0.16\linewidth]{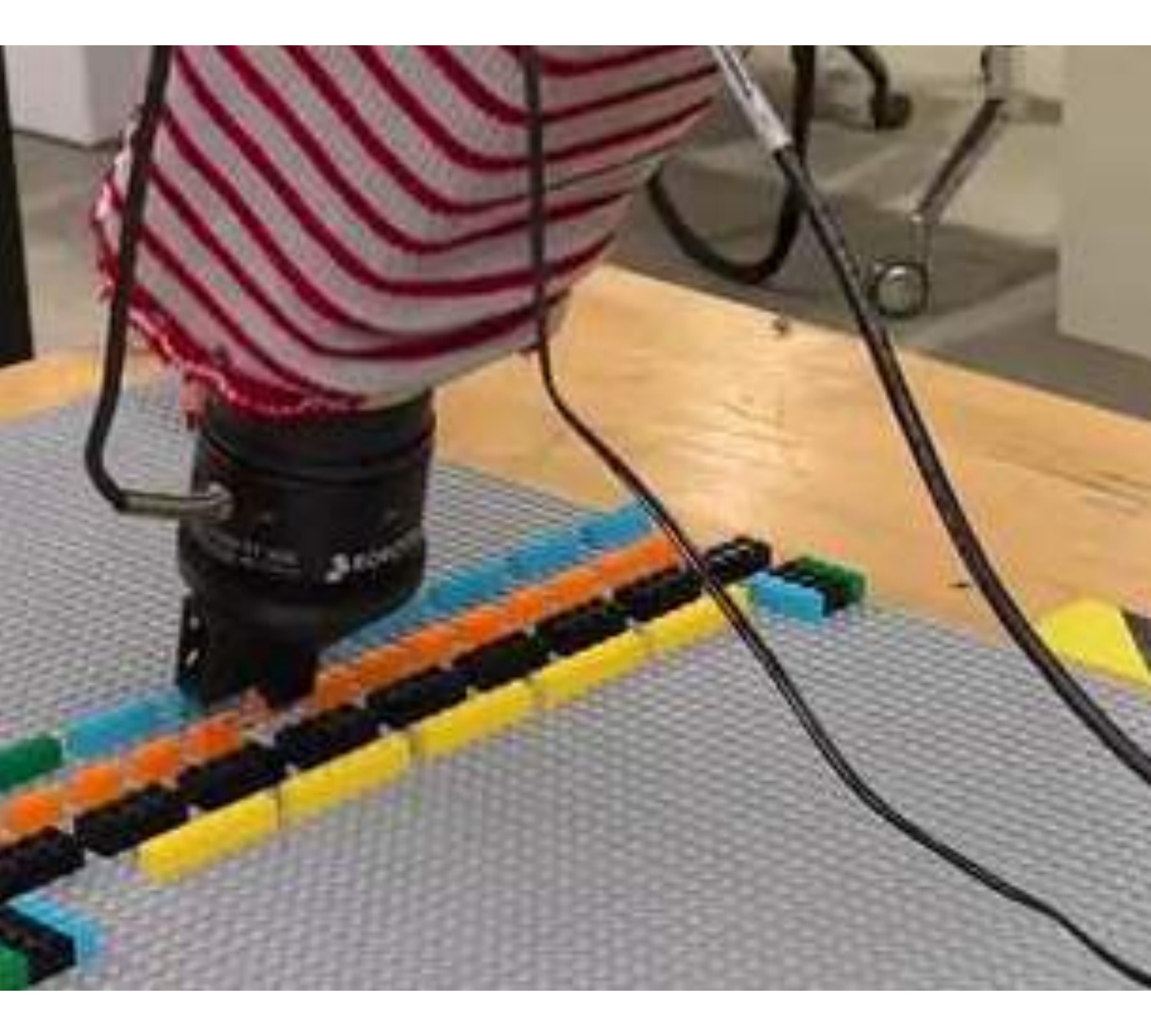}\label{fig:7}}\hfill
\subfigure[]{\includegraphics[width=0.16\linewidth]{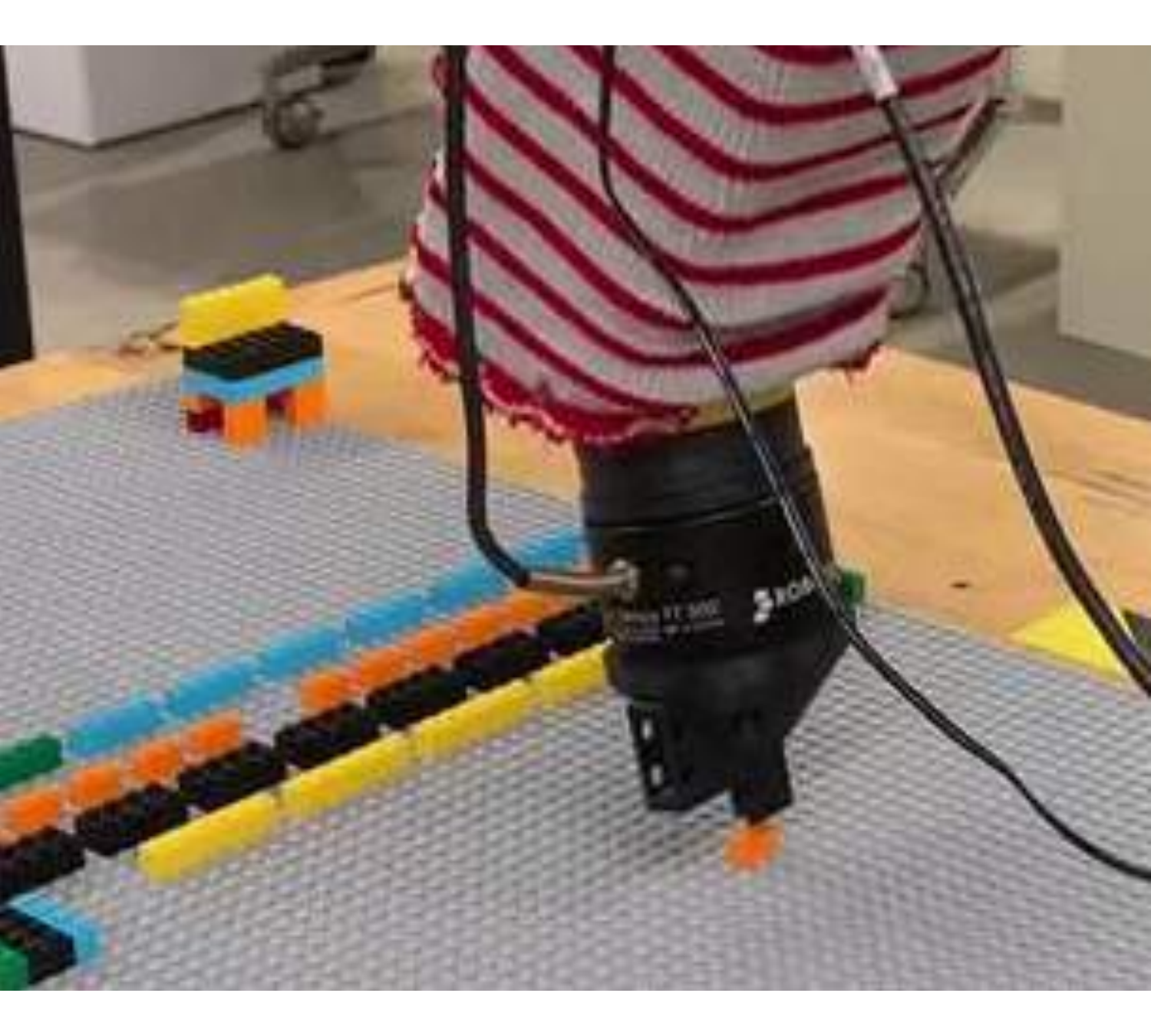}\label{fig:8}}\hfill
\subfigure[]{\includegraphics[width=0.16\linewidth]{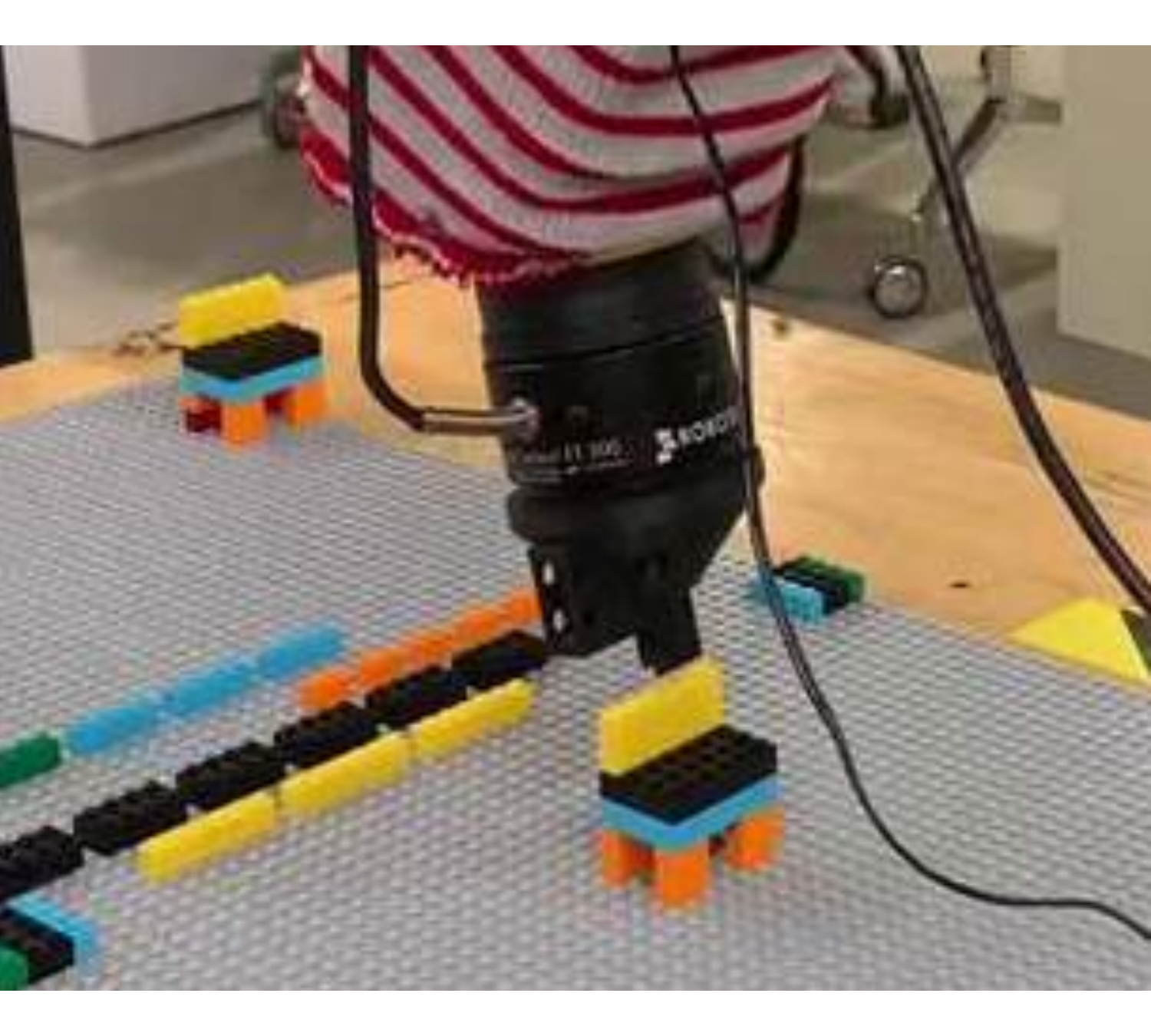}\label{fig:9}}\hfill
\subfigure[]{\includegraphics[width=0.16\linewidth]{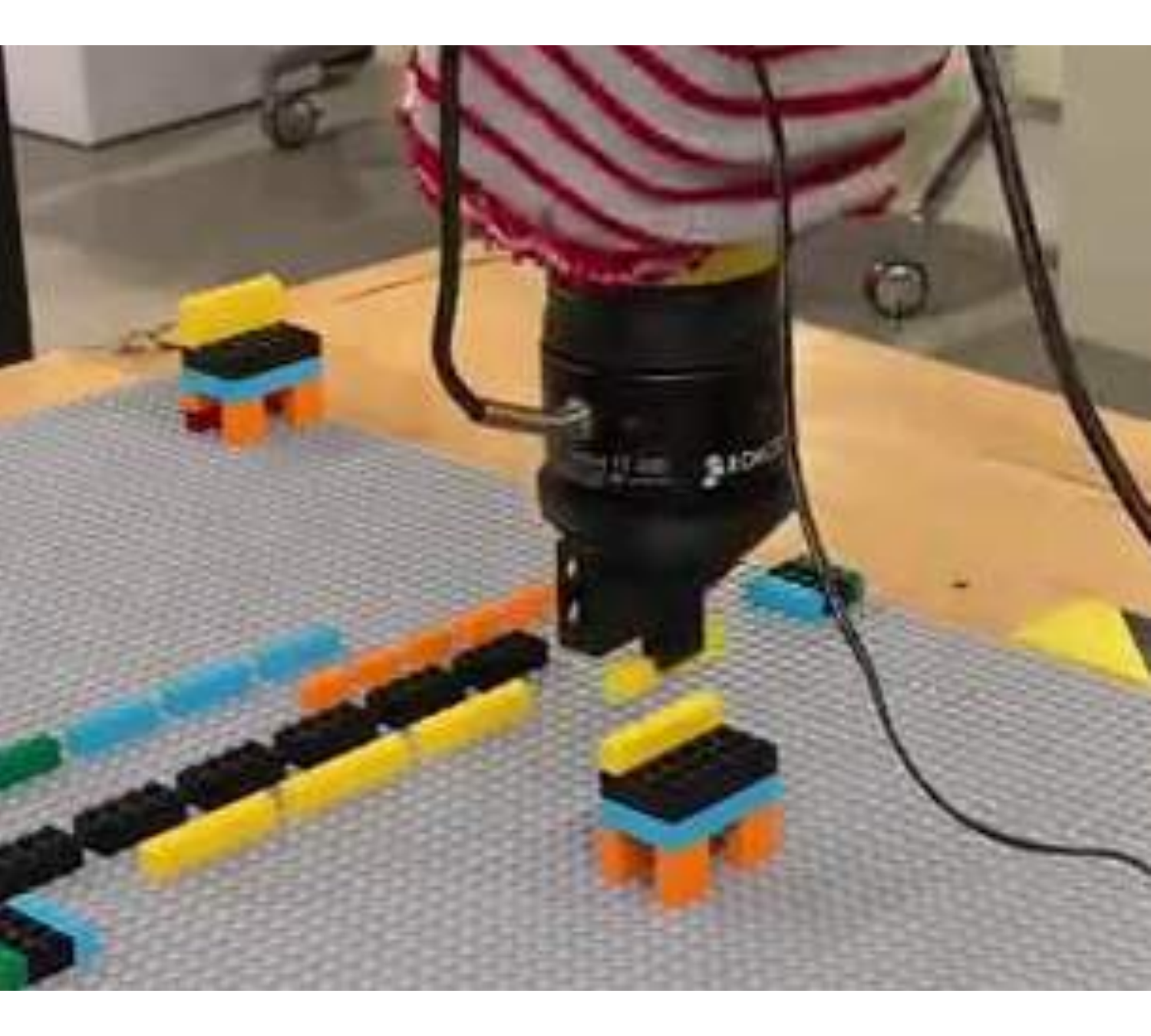}\label{fig:10}}\hfill
\subfigure[]{\includegraphics[width=0.16\linewidth]{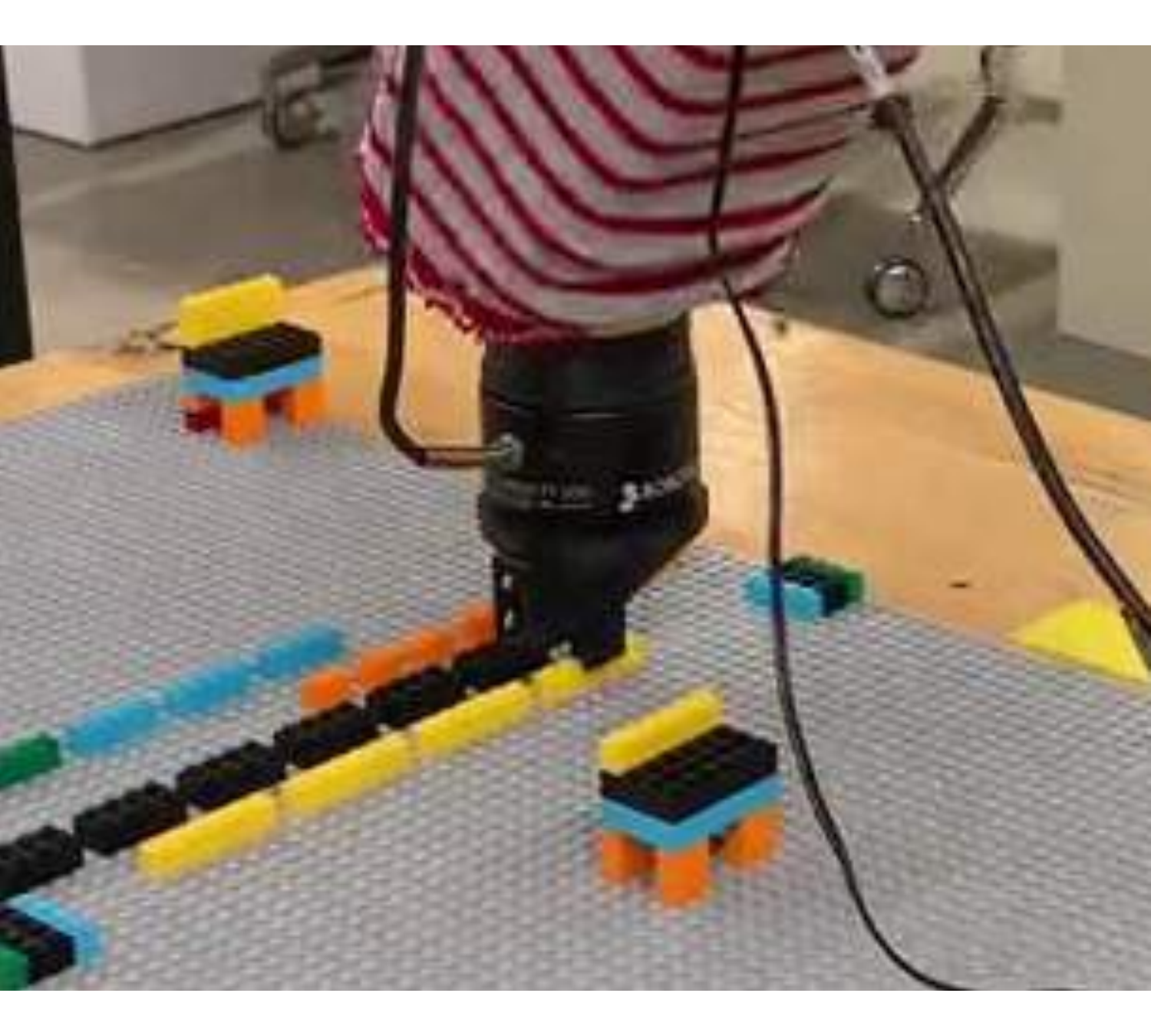}\label{fig:11}}\hfill
\subfigure[]{\includegraphics[width=0.16\linewidth]{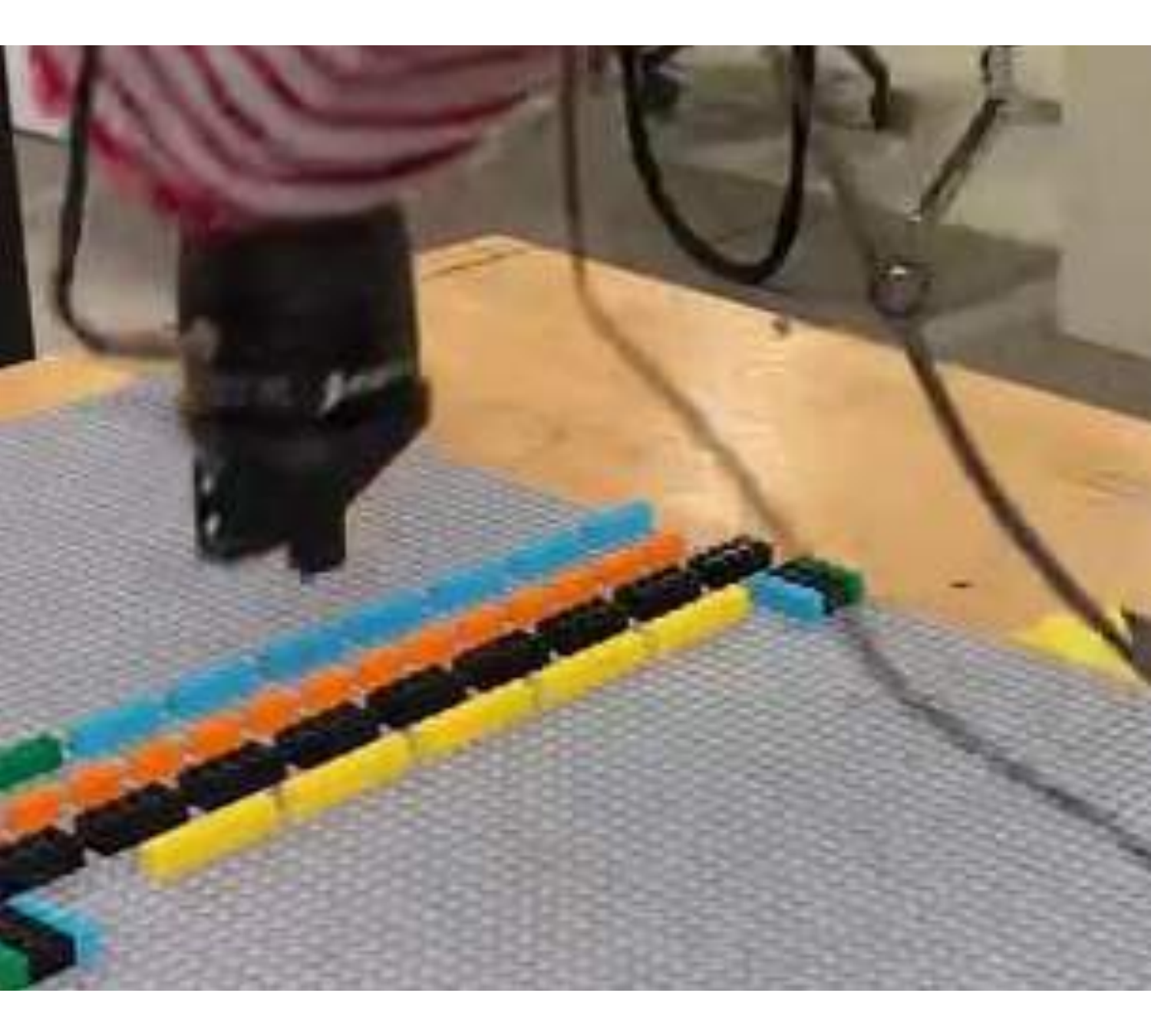}\label{fig:12}}
\\
\subfigure[]{\includegraphics[width=0.16\linewidth]{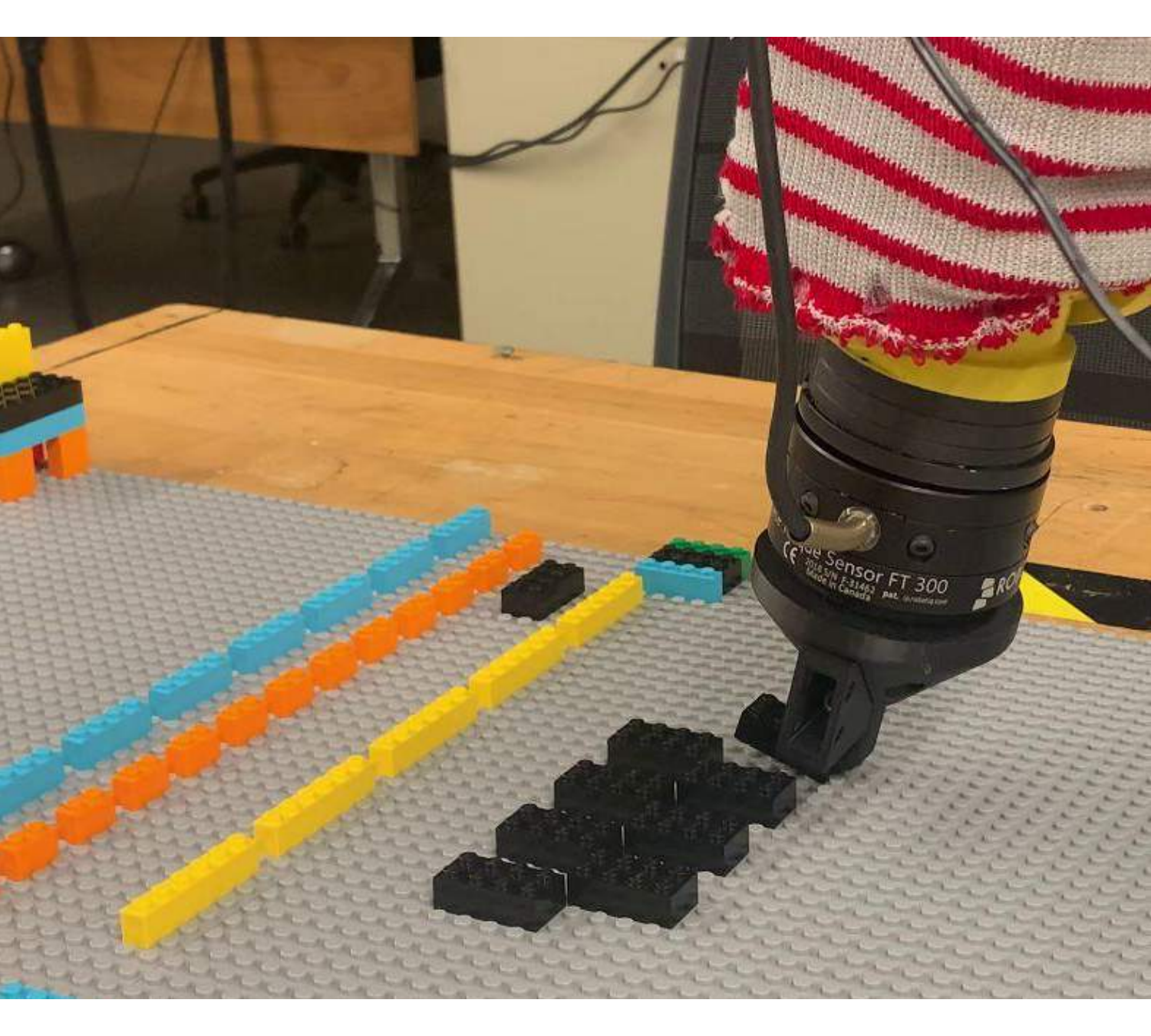}\label{fig:13}}\hfill
\subfigure[]{\includegraphics[width=0.16\linewidth]{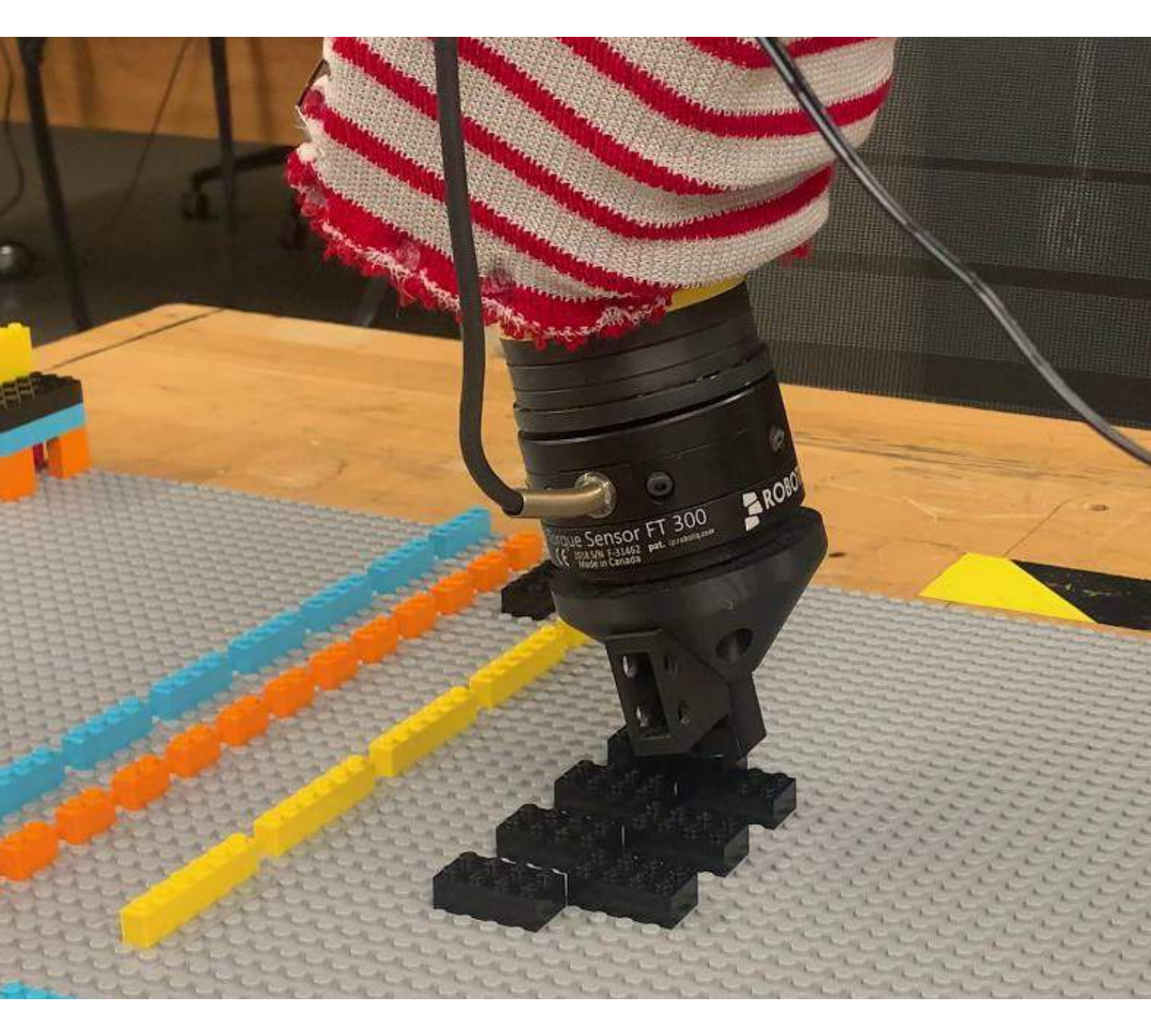}\label{fig:14}}\hfill
\subfigure[]{\includegraphics[width=0.16\linewidth]{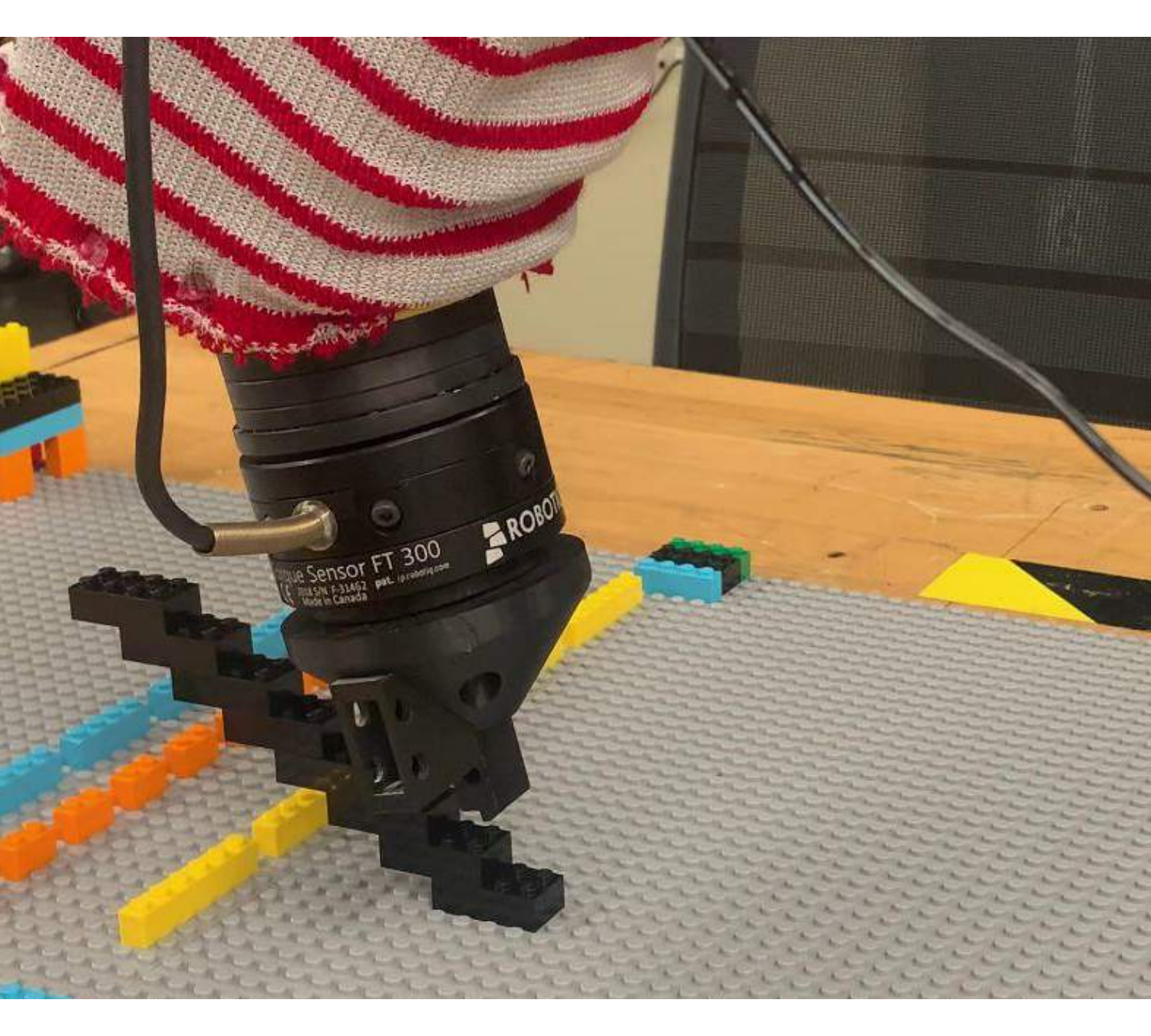}\label{fig:15}}\hfill
\subfigure[]{\includegraphics[width=0.16\linewidth]{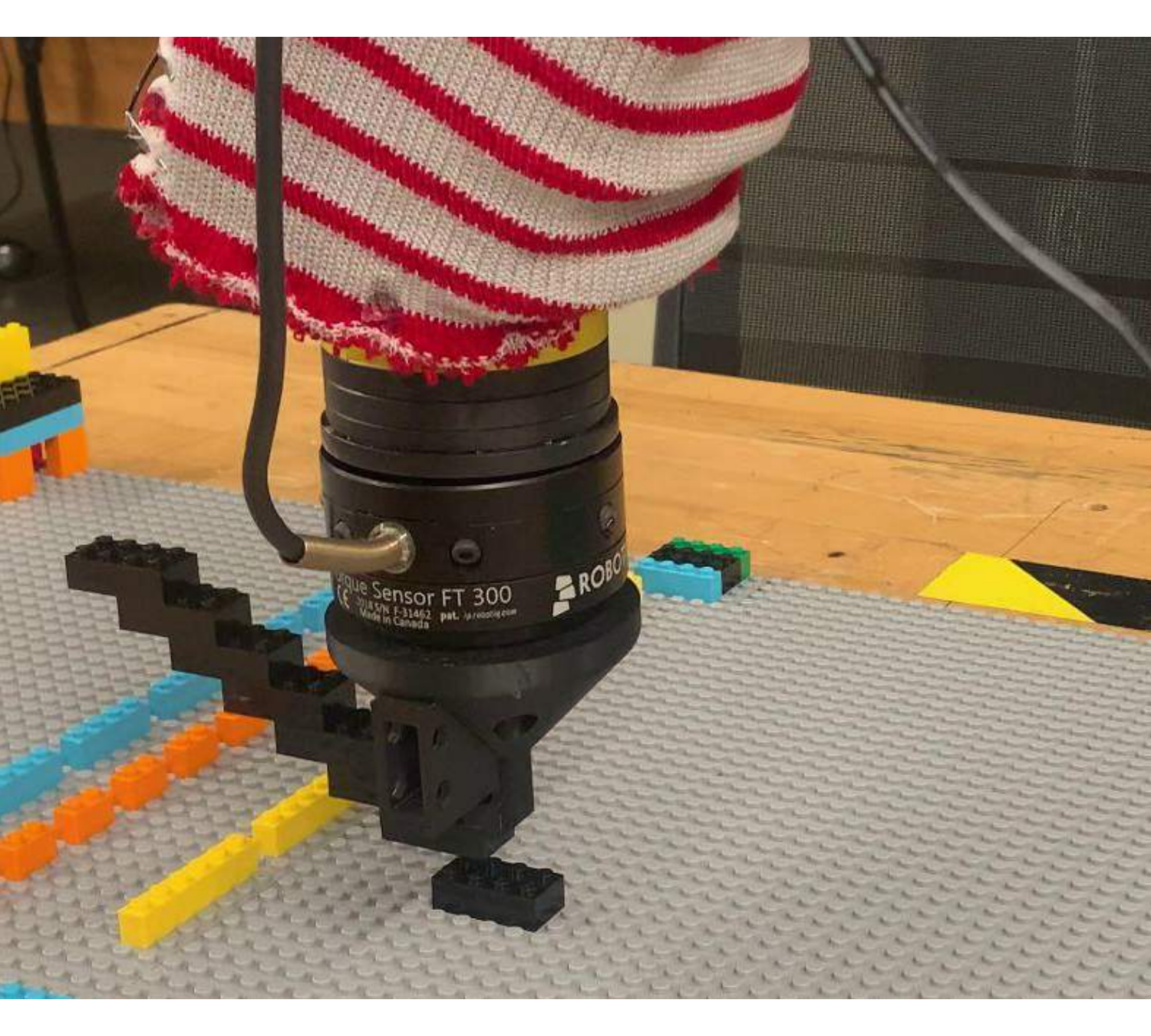}\label{fig:16}}\hfill
\subfigure[]{\includegraphics[width=0.16\linewidth]{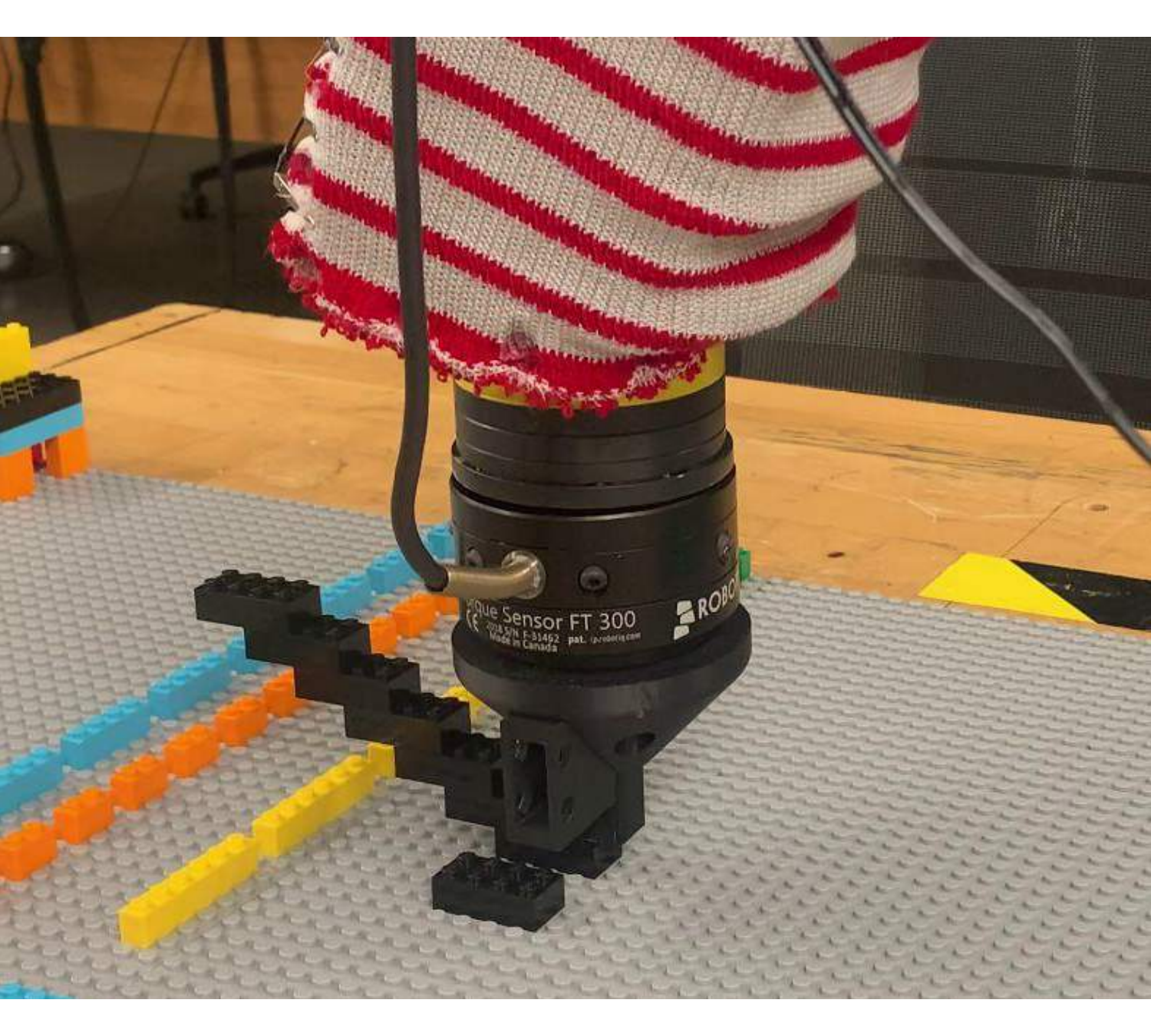}\label{fig:17}}\hfill
\subfigure[]{\includegraphics[width=0.16\linewidth]{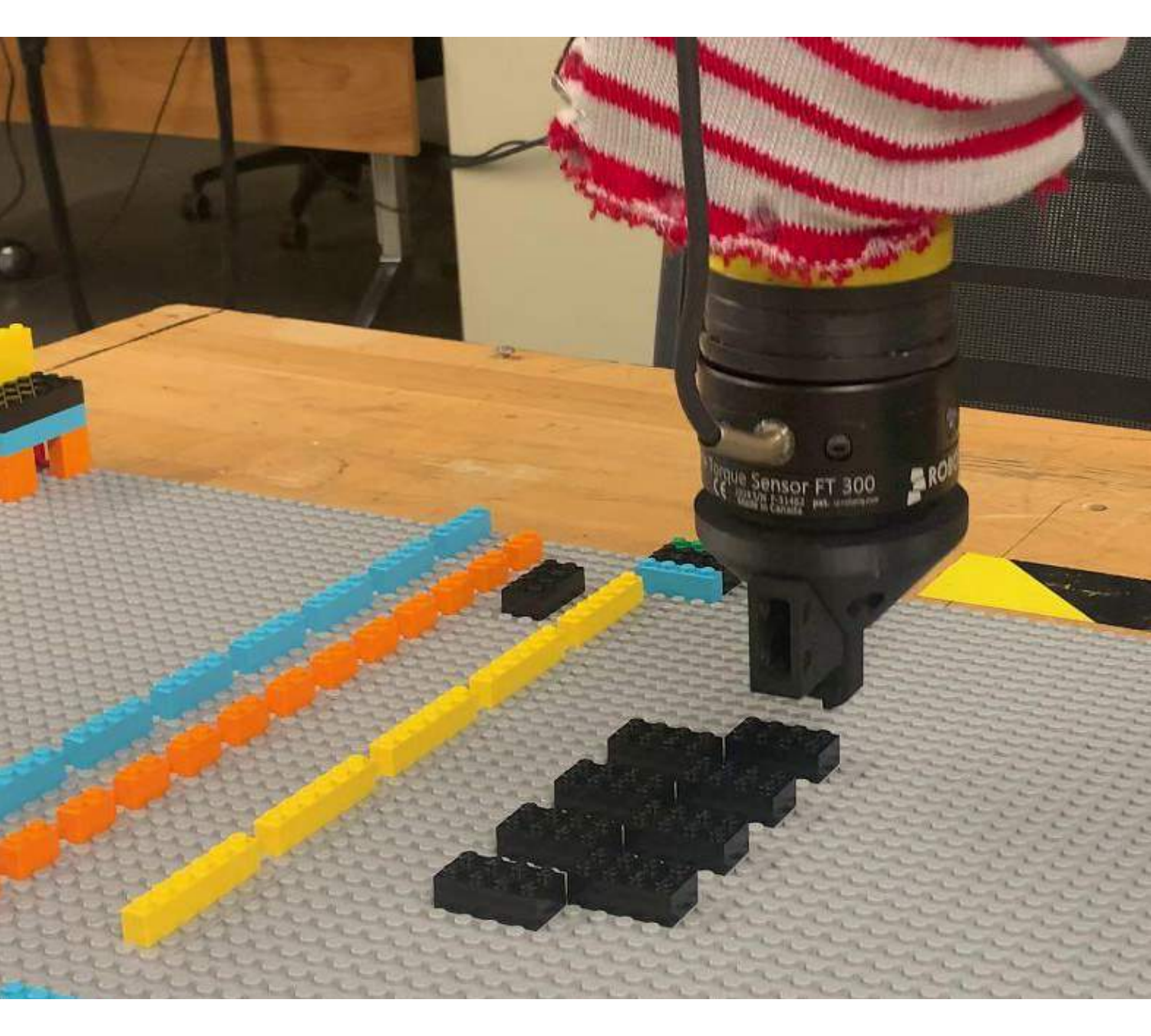}\label{fig:18}}\\
    \caption{Sustainable Rapid Lego Prototyping. Top: RI (2D). Middle: A chair (3D). Bottom: Stairs (3D).} \label{fig:lego_prototyping}
\end{figure*}

To improve the manipulation performance, we apply \cref{alg:cmaes} to optimize the manipulation parameters. 
We compare the manipulation performances with the initial parameters 
and optimized parameters.
The safe robot learning is done on 8 sampled positions on the plate with 1 layer of the $1\times 2$ brick.
The bounds are set to $\theta\in [1^\circ,25^\circ]$, $d_x,d_z\in [0, 10]\si{mm}$ based on empirical observations. We set $\infty$ as $10^8$ and $\sigma=0.005$ in the implementation.
The robot interface requires the received trajectory to have bounded jerk in the joint space.
Since the joint JPC inherently ensures control limit, we have $\alpha=100$, $\beta=10$, $\gamma=100$, and $\eta=1$.
However, the Cartesian JPC only ensures the control limit in the Carteisan-space, which might violate the joint-space control bound when inverse mapping to the joint space. 
Therefore, we have $\alpha=100$, $\beta=10$, $\gamma=100$, and $\eta=100$ to smooth the robot control profile.
The top figures in \Cref{fig:learning_costs} display the evolution of the costs for different tasks with different controllers.
The proposed learning technique effectively reduces the cost when solving \cref{eq:obj} and the learned parameters are converging.
Note that the cost value might converge slower than the parameter values since \cref{eq:cost} is a piece-wise value function. 
It can be seen that although the parameters are converging, the cost is not mainly because the decreasing $T$ and $\theta$ push the performance to the boundary.
Therefore, slight changes in the parameters would easily cause the manipulation to fail, making the cost value fluctuate significantly.
As the parameters further converge and sample variances decrease, the cost value then converges.
Also note that after learning, the execution time for joint JPC significantly decreases (\ie from 2\si{s} to 0.4\si{s} for assembly and 0.6\si{s} for disassembly in the left of \cref{fig:learning_costs}), but the time for Cartesian JPC increases (\ie from 2\si{s} to 5.9\si{s} for assembly and 6.1\si{s} for disassembly in the right of \cref{fig:learning_costs}). 
This is because a shorter execution time leads to a more aggressive control profile. 
Thus, due to the large $\eta$, the execution time increases to ensure a smooth control profile.
\Cref{table:EOAT_performance} compares the manipulation performance using initial and optimized manipulation parameters.
The performance improves when we optimize the manipulation parameters as the robot can manipulate Lego bricks at a 100\% success rate.
The faster execution and higher success rate demonstrate that the robot can achieve fast and safe Lego manipulation.
Moreover, the results demonstrate that the learning framework is controller independent, and the parameters learned from a small sample space (\ie 1 layer of $1\times 2$ brick at 8 positions) are generalizable to different bricks with different structures at different positions.
This is because the designed EOAT simplifies the manipulation and unifies different tasks using the parameters $\theta, d_x, d_z$.

\paragraph{Transferability to Different Robot Platforms}
To illustrate the transferability, we deploy the proposed hardware-software co-design to Yaskawa GP4 robots as shown in \cref{fig:yaskawa}.
We use ATI Gamma FTS on the Yaskawa robots.
With the proposed EOAT, the Yaskawa robots are capable of robustly manipulating Lego bricks without further tuning. 
Because the designed EOAT reduces the manipulation space to only $\theta$, $d_x$, and $d_z$, the optimized manipulation capability is transferable to different robots even though it is optimized on another platform (\ie FANUC robot).

\paragraph{Sustainable Robotic Lego Prototyping}
\label{sec:robotic_prototype}
The experiments demonstrate that with the designed EOAT and safe robot learning, the robot is capable of rapidly and reliably manipulating Lego bricks.
The manipulation capability enables sustainable robotic prototyping. 
\Cref{fig:lego_prototyping} demonstrates the robot building Lego prototypes and breaking them to restore the workspace.
\Cref{fig:1,fig:2,fig:3,fig:4,fig:5,fig:6} illustrate the robot prototyping two characters: \textit{RI}.
In \cref{fig:7,fig:8,fig:9,fig:10,fig:11,fig:12}, the robot prototypes a 3D chair.
And the robot accomplishes 3D stairs in \cref{fig:13,fig:14,fig:15,fig:16,fig:17,fig:18}.
The robot builds the Lego prototype by disassembling bricks from the storage plate and assembling them on the working plate.
It restores the environment by disassembling bricks from the workspace and assembling them in the storage space.
With the capability of manipulating Lego bricks, we can achieve a sustainable robotic Lego prototyping system, which is able to build, restore, and restart new tasks without human intervention.

\section{DISCUSSION}
\label{sec:limitations}
\paragraph{Limitations}

The current framework directly optimizes parameters in the real-world environment. 
Therefore, a human operator is needed to observe whether the robot succeeded in the manipulation task to calculate \cref{eq:cost}.
In addition, the human is required to reset the environment after each trial.
As a result, the human effort would slow down the learning process.
In addition, as discussed in \cref{sec:result}, the Lego plates are pre-calibrated, and the robot operates in an open loop.
Given a well-calibrated environment, the robot can robustly manipulate Lego bricks.
However, the performance would be influenced by the calibration quality.

\paragraph{Future works}
In the future, we aim to alleviate the required human effort and accelerate safe learning.
In particular, an external vision sensor can be used to determine whether the manipulation is successful. 
In addition, we can replace the human operator with another robot, which can restore the environment if the manipulation fails during learning.
To improve the prototyping system, we aim to integrate a feedback loop (\eg real-time force feedback) to improve the prototyping robustness. Moreover, we aim to open source the EOAT design and provide an open platform for Lego manipulation and prototyping systems.


\section{CONCLUSION}
\label{sec:conclusion}

This paper studies safe and efficient Lego manipulation and presents a robotic solution that can assemble and disassemble Lego bricks. 
In particular, an EOAT is designed, which allows robots to manipulate Lego bricks.
In addition, this paper uses evolution strategy to safely optimize the robot motion for Lego manipulation.
The system is deployed to FANUC LR-mate 200id/7L and Yaskawa GP4 robots.
Experiments demonstrate that the EOAT performs reliably in manipulating Lego bricks and the learning framework can effectively and safely improve the manipulation performance to a 100\% success rate.
Moreover, the manipulation capability is transferable and it can enable rapid sustainable robotic prototyping in the future.

\section*{ACKNOWLEDGMENT}
The authors would like to thank Mr. Shobhit Aggarwal and the Manufacturing Futures Institute, Carnegie Mellon University, for setting up the Yaskawa GP4 robots.

\bibliographystyle{IEEEtran}
\bibliography{IEEEexample}

\begin{thebibliography}{10}
\providecommand{\url}[1]{#1}
\csname url@rmstyle\endcsname
\providecommand{\newblock}{\relax}
\providecommand{\bibinfo}[2]{#2}
\providecommand\BIBentrySTDinterwordspacing{\spaceskip=0pt\relax}
\providecommand\BIBentryALTinterwordstretchfactor{4}
\providecommand\BIBentryALTinterwordspacing{\spaceskip=\fontdimen2\font plus
\BIBentryALTinterwordstretchfactor\fontdimen3\font minus \fontdimen4\font\relax}
\providecommand\BIBforeignlanguage[2]{{%
\expandafter\ifx\csname l@#1\endcsname\relax
\typeout{** WARNING: IEEEtran.bst: No hyphenation pattern has been}%
\typeout{** loaded for the language `#1'. Using the pattern for}%
\typeout{** the default language instead.}%
\else
\language=\csname l@#1\endcsname
\fi
#2}}

\bibitem{10.1007/978-3-319-33609-1_2}
A.~Khan and K.~Turowski, ``A survey of current challenges in manufacturing industry and preparation for industry 4.0,'' in \emph{Proceedings of the First International Scientific Conference ``Intelligent Information Technologies for Industry'' (IITI'16)}, A.~Abraham, S.~Kovalev, V.~Tarassov, and V.~Sn{\'a}{\v{s}}el, Eds.\hskip 1em plus 0.5em minus 0.4em\relax Cham: Springer International Publishing, 2016, pp. 15--26.

\bibitem{review_additive}
K.~Wong, ``K.v. wong, a.hernandez, “a review of additive manufacturing,” isrn mechanical engineering, vol 2012 (2012), article id 208760, 10 pages.'' \emph{ISRN Mechanical Engineering}, vol. 2012, 08 2012.

\bibitem{RASIYA20216896}
G.~Rasiya, A.~Shukla, and K.~Saran, ``Additive manufacturing-a review,'' \emph{Materials Today: Proceedings}, vol.~47, pp. 6896--6901, 2021, international Conference on Advances in Design, Materials and Manufacturing.

\bibitem{10.5555/3113190.3113359}
A.~Seth, J.~M. Vance, and J.~H. Oliver, ``Virtual reality for assembly methods prototyping: A review,'' \emph{Virtual Real.}, vol.~15, no.~1, p. 5–20, mar 2011.

\bibitem{AHMAD2015412}
A.~AHMAD, S.~DARMOUL, W.~AMEEN, M.~H. ABIDI, and A.~M. AL-AHMARI, ``Rapid prototyping for assembly training and validation,'' \emph{IFAC-PapersOnLine}, vol.~48, no.~3, pp. 412--417, 2015, 15th IFAC Symposium onInformation Control Problems inManufacturing.

\bibitem{GOMESDESA1999389}
A.~{Gomes de Sá} and G.~Zachmann, ``Virtual reality as a tool for verification of assembly and maintenance processes,'' \emph{Computers \& Graphics}, vol.~23, no.~3, pp. 389--403, 1999.

\bibitem{9817380}
M.-L. Lee, W.~Liu, S.~Behdad, X.~Liang, and M.~Zheng, ``Robot-assisted disassembly sequence planning with real-time human motion prediction,'' \emph{IEEE Transactions on Systems, Man, and Cybernetics: Systems}, vol.~53, no.~1, pp. 438--450, 2023.

\bibitem{disassemble_article}
K.~K. Pochampally, \emph{European Journal of Operational Research}, vol. 187, no.~1, pp. 335--337, 2008.

\bibitem{lego_spike_prime}
I.~H. Pérez~Tavera, ``Lego education - spike prime,'' \emph{Vida Científica Boletín Científico de la Escuela Preparatoria No. 4}, vol.~10, no.~19, pp. 9--11, ene. 2022.

\bibitem{doi:10.5772/58249}
E.~Danahy, E.~Wang, J.~Brockman, A.~Carberry, B.~Shapiro, and C.~B. Rogers, ``Lego-based robotics in higher education: 15 years of student creativity,'' \emph{International Journal of Advanced Robotic Systems}, vol.~11, no.~2, p.~27, 2014.

\bibitem{ZHOU2020103282}
C.~Zhou, B.~Tang, L.~Ding, P.~Sekula, Y.~Zhou, and Z.~Zhang, ``Design and automated assembly of planetary lego brick for lunar in-situ construction,'' \emph{Automation in Construction}, vol. 118, p. 103282, 2020.

\bibitem{Kim2014SurveyOA}
J.~W. Kim, ``Survey on automated lego assembly construction,'' 2014.

\bibitem{10.1145/2816795.2818091}
S.-J. Luo, Y.~Yue, C.-K. Huang, Y.-H. Chung, S.~Imai, T.~Nishita, and B.-Y. Chen, ``Legolization: Optimizing lego designs,'' \emph{ACM Trans. Graph.}, vol.~34, no.~6, nov 2015.

\bibitem{doi:10.1177/09544054211053616}
B.~Zhou, T.~Tian, J.~Zhao, and D.~Liu, ``A legorization method based on 3d color printing trajectory,'' \emph{Proceedings of the Institution of Mechanical Engineers, Part B: Journal of Engineering Manufacture}, vol. 236, no. 6-7, pp. 844--867, 2022.

\bibitem{10.1111:cgf.13603}
J.~Zhou, X.~Chen, and Y.~Xu, ``{Automatic Generation of Vivid LEGO Architectural Sculptures},'' \emph{Computer Graphics Forum}, 2019.

\bibitem{thompson2020Lego}
R.~Thompson, G.~Elahe, T.~DeVries, and G.~W. Taylor, ``Building lego using deep generative models of graphs,'' \emph{Machine Learning for Engineering Modeling, Simulation, and Design Workshop at Neural Information Processing Systems}, 2020.

\bibitem{10.1007/978-3-031-19815-1_6}
A.~Walsman, M.~Zhang, K.~Kotar, K.~Desingh, A.~Farhadi, and D.~Fox, ``Break and make: Interactive structural understanding using lego bricks,'' in \emph{Computer Vision -- ECCV 2022}, S.~Avidan, G.~Brostow, M.~Ciss{\'e}, G.~M. Farinella, and T.~Hassner, Eds.\hskip 1em plus 0.5em minus 0.4em\relax Cham: Springer Nature Switzerland, 2022, pp. 90--107.

\bibitem{wang2022translating}
R.~Wang, Y.~Zhang, J.~Mao, C.-Y. Cheng, and J.~Wu, ``Translating a visual lego manual to a machine-executable plan,'' in \emph{European Conference on Computer Vision}, 2022.

\bibitem{LegoBuilder}
S.~Ono, A.~Andre, Y.~Chang, and M.~Nakajima, ``Lego builder: Automatic generation of lego assembly manual from 3d polygon model,'' \emph{ITE Transactions on Media Technology and Applications}, vol.~1, pp. 354--360, 10 2013.

\bibitem{ChungH2021neurips}
H.~Chung, J.~Kim, B.~Knyazev, J.~Lee, G.~W. Taylor, J.~Park, and M.~Cho, ``{Brick-by-Brick}: Combinatorial construction with deep reinforcement learning,'' in \emph{Advances in Neural Information Processing Systems (NeurIPS)}, vol.~34, 2021.

\bibitem{KimJ2020arxiv}
J.~Kim, H.~Chung, J.~Lee, M.~Cho, and J.~Park, ``Combinatorial {3D} shape generation via sequential assembly,'' \emph{{arXiv} preprint {arXiv}:2004.07414}, 2020.

\bibitem{lennon2021image2lego}
K.~Lennon, K.~Fransen, A.~O'Brien, Y.~Cao, M.~Beveridge, Y.~Arefeen, N.~Singh, and I.~Drori, ``Image2lego: Customized lego set generation from images,'' \emph{{arXiv} preprint {arXiv}:2108.08477}, 2021.

\bibitem{ahn2022sequential}
S.~Ahn, J.~Kim, M.~Cho, and J.~Park, ``Sequential brick assembly with efficient constraint satisfaction,'' \emph{{arXiv} preprint {arXiv}:2210.01021}, 2022.

\bibitem{8419684}
S.-M. Lee, J.~W. Kim, and H.~Myung, ``Split-and-merge-based genetic algorithm (sm-ga) for lego brick sculpture optimization,'' \emph{IEEE Access}, vol.~6, pp. 40\,429--40\,438, 2018.

\bibitem{popov2017dataefficient}
I.~Popov, N.~Heess, T.~Lillicrap, R.~Hafner, G.~Barth-Maron, M.~Vecerik, T.~Lampe, Y.~Tassa, T.~Erez, and M.~Riedmiller, ``Data-efficient deep reinforcement learning for dexterous manipulation,'' \emph{arXiv}, 2017.

\bibitem{10.1109/ICRA.2019.8793659}
Y.~Fan, J.~Luo, and M.~Tomizuka, ``A learning framework for high precision industrial assembly,'' in \emph{2019 International Conference on Robotics and Automation (ICRA)}.\hskip 1em plus 0.5em minus 0.4em\relax IEEE Press, 2019, p. 811–817.

\bibitem{8674203}
Z.~Zhu, H.~Hu, and D.~Gu, ``Robot performing peg-in-hole operations by learning from human demonstration,'' in \emph{2018 10th Computer Science and Electronic Engineering (CEEC)}, 2018, pp. 30--35.

\bibitem{9341428}
L.~Nägele, A.~Hoffmann, A.~Schierl, and W.~Reif, ``Legobot: Automated planning for coordinated multi-robot assembly of lego structures,'' in \emph{2020 IEEE/RSJ International Conference on Intelligent Robots and Systems (IROS)}, 2020, pp. 9088--9095.

\bibitem{7759340}
Y.~Maeda, O.~Nakano, T.~Maekawa, and S.~Maruo, ``From cad models to toy brick sculptures: A 3d block printer,'' in \emph{2016 IEEE/RSJ International Conference on Intelligent Robots and Systems (IROS)}, 2016, pp. 2167--2172.

\bibitem{9812161}
G.~Zhou, L.~Luo, H.~Xu, X.~Zhang, H.~Guo, and H.~Zhao, ``Brick yourself within 3 minutes,'' in \emph{2022 International Conference on Robotics and Automation (ICRA)}, 2022, pp. 6261--6267.

\bibitem{8593852}
K.~Gilday, J.~Hughes, and F.~Iida, ``Achieving flexible assembly using autonomous robotic systems,'' in \emph{2018 IEEE/RSJ International Conference on Intelligent Robots and Systems (IROS)}, 2018, pp. 1--9.

\bibitem{stablebaselines3}
\BIBentryALTinterwordspacing
A.~Raffin, A.~Hill, A.~Gleave, A.~Kanervisto, M.~Ernestus, and N.~Dormann, ``Stable-baselines3: Reliable reinforcement learning implementations,'' \emph{Journal of Machine Learning Research}, vol.~22, no. 268, pp. 1--8, 2021. [Online]. Available: \url{http://jmlr.org/papers/v22/20-1364.html}
\BIBentrySTDinterwordspacing

\bibitem{zhao2023guard}
W.~Zhao, R.~Chen, Y.~Sun, R.~Liu, T.~Wei, and C.~Liu, ``Guard: A safe reinforcement learning benchmark,'' \emph{arXiv preprint arXiv:2305.13681}, 2023.

\bibitem{liu2023robotic}
R.~Liu, Y.~Sun, and C.~Liu, ``Robotic lego assembly and disassembly from human demonstration,'' \emph{arXiv preprint arXiv:2305.15667}, 2023.

\bibitem{liu2023simulation}
R.~Liu, A.~Chen, X.~Luo, and C.~Liu, ``Simulation-aided learning from demonstration for robotic lego construction,'' \emph{arXiv preprint arXiv:2309.11010}, 2023.

\bibitem{zhao2024autonomous}
Y.~Zhao and T.~Kato, ``Autonomous robust assembly planning,'' Mar.~26 2024, uS Patent 11,938,633.

\bibitem{hansen2016cma}
N.~Hansen, ``The cma evolution strategy: A tutorial,'' \emph{arXiv preprint arXiv:1604.00772}, 2016.

\bibitem{nomura2021warm}
M.~Nomura, S.~Watanabe, Y.~Akimoto, Y.~Ozaki, and M.~Onishi, ``Warm starting cma-es for hyperparameter optimization,'' in \emph{Proceedings of the AAAI Conference on Artificial Intelligence}, vol.~35, no.~10, 2021, pp. 9188--9196.

\bibitem{chen2022composable}
R.~Chen, C.~Wang, T.~Wei, and C.~Liu, ``A composable framework for policy design, learning, and transfer toward safe and efficient industrial insertion,'' \emph{{arXiv} preprint {arXiv}:2203.03017}, 2022.

\bibitem{9029720}
T.~Wei and C.~Liu, ``Safe control algorithms using energy functions: A uni ed framework, benchmark, and new directions,'' in \emph{2019 IEEE 58th Conference on Decision and Control (CDC)}, 2019, pp. 238--243.

\bibitem{jpc}
R.~Liu, R.~Chen, Y.~Sun, Y.~Zhao, and C.~Liu, ``Jerk-bounded position controller with real-time task modification for interactive industrial robots,'' in \emph{2022 IEEE/ASME International Conference on Advanced Intelligent Mechatronics (AIM)}, 2022, pp. 1771--1778.

\bibitem{jssa}
R.~Liu, R.~Chen, and C.~Liu, ``Safe interactive industrial robots using jerk-based safe set algorithm,'' \emph{Proceedings of the International Symposium on Flexible Automation}, vol. 2022, pp. 196--203, 2022.

\bibitem{ssa}
C.~Liu and M.~Tomizuka, ``Control in a safe set: Addressing safety in human-robot interactions,'' in \emph{ASME Dynamic Systems and Control Conference}, vol.~3, 11 2014.

\bibitem{1087247}
O.~Khatib, ``Real-time obstacle avoidance for manipulators and mobile robots,'' in \emph{Proceedings. 1985 IEEE International Conference on Robotics and Automation}, vol.~2, 1985, pp. 500--505.

\bibitem{6414600}
L.~Gracia, F.~Garelli, and A.~Sala, ``Reactive sliding-mode algorithm for collision avoidance in robotic systems,'' \emph{IEEE Transactions on Control Systems Technology}, vol.~21, no.~6, pp. 2391--2399, 2013.

\bibitem{7040372}
A.~D. Ames, J.~W. Grizzle, and P.~Tabuada, ``Control barrier function based quadratic programs with application to adaptive cruise control,'' in \emph{53rd IEEE Conference on Decision and Control}, 2014, pp. 6271--6278.

\bibitem{taskagnostic}
\BIBentryALTinterwordspacing
R.~Liu, R.~Chen, and C.~Liu, ``Task-agnostic adaptation for safe human-robot handover,'' \emph{IFAC-PapersOnLine}, vol.~55, no.~41, pp. 175--180, 2022, 4th IFAC Workshop on Cyber-Physical and Human Systems CPHS 2022. [Online]. Available: \url{https://www.sciencedirect.com/science/article/pii/S2405896323001295}
\BIBentrySTDinterwordspacing

\bibitem{10252579}
R.~Liu, R.~Chen, A.~Abuduweili, and C.~Liu, ``Proactive human-robot co-assembly: Leveraging human intention prediction and robust safe control,'' in \emph{2023 IEEE Conference on Control Technology and Applications (CCTA)}, 2023, pp. 339--345.

\bibitem{wei2023zero}
T.~Wei, S.~Kang, R.~Liu, and C.~Liu, ``Zero-shot transferable and persistently feasible safe control for high dimensional systems by consistent abstraction,'' in \emph{2023 62nd IEEE Conference on Decision and Control (CDC)}, 2023, pp. 8614--8619.

\bibitem{chen2024real}
R.~Chen, W.~Zhao, R.~Liu, W.~Zhang, and C.~Liu, ``Real-time safety index adaptation for parameter-varying systems via determinant gradient ascend,'' \emph{arXiv preprint arXiv:2403.14968}, 2024.

\end{thebibliography}

\end{document}